\documentclass[lettersize,journal]{IEEEtran}

\usepackage[switch]{lineno}
\usepackage{amsmath,amsfonts}
\allowdisplaybreaks
\usepackage{algorithm}
\usepackage{array}
\usepackage{textcomp}
\usepackage{stfloats}
\usepackage{url}
\usepackage{verbatim}
\usepackage{graphicx}
\usepackage{cite}
\usepackage{subcaption}
\usepackage{xspace}
\usepackage{paralist}
\usepackage{enumitem}
\usepackage{wrapfig}
\usepackage[noend]{algpseudocode}
\usepackage{threeparttable}
\usepackage{booktabs}
\usepackage{multirow}
\usepackage{xcolor}
\usepackage[table]{xcolor}

\definecolor{mygreen}{RGB}{200, 255, 200}
\newcommand{\tbgreen}{\cellcolor{mygreen}}

\newtheorem{definition}{Definition}

\newtheorem{example}{Example}

\newcounter{researchquestionCount}

\newcommand{\researchquestion}[1]{%
  \refstepcounter{researchquestionCount}%
  \medskip\noindent
  \textbf{RQ\arabic{researchquestionCount}}\quad\textit{#1}%
  \par\smallskip
}

\newcommand{\reluStateOne}{\ensuremath{\mathcal{I}}\xspace}
\newcommand{\reluStateTwo}{\ensuremath{\widehat{\mathcal{I}}}\xspace}
\newcommand{\reluStateRel}{\ensuremath{\mathcal{I}^{\Delta}}\xspace}
\newcommand{\reluNeg}{\mathcal{I}^{-}}
\newcommand{\reluNegHat}{\widehat{\mathcal{I}}^{-}}
\newcommand{\reluPos}{\mathcal{I}^{+}}
\newcommand{\reluPosHat}{\widehat{\mathcal{I}}^{+}}
\newcommand{\reluUns}{\mathcal{I}^{\pm}}
\newcommand{\reluUnsHat}{\widehat{\mathcal{I}}^{\pm}}
\newcommand{\relReluNeg}{\mathcal{I}^{\Delta-}}
\newcommand{\relReluPos}{\mathcal{I}^{\Delta+}}
\newcommand{\relReluUns}{\mathcal{I}^{\Delta\pm}}

\newcommand{\inpNeg}{\mathcal{E^{-}}}
\newcommand{\inpPos}{\mathcal{E^{+}}}
\newcommand{\inpHatNeg}{{\widehat{\mathcal{E}}}^{-}}
\newcommand{\inpHatPos}{{\widehat{\mathcal{E}}}^{+}}

\newcommand{\inpDeltaNeg}{\mathcal{E}^{{\Delta-}}}
\newcommand{\inpDeltaPos}{\mathcal{E}^{{\Delta+}}}

\NewDocumentCommand{\affineInd}{o}{
  \IfNoValueTF{#1}
    {\mathcal{A}}
    {\mathcal{A}^{(#1)}}}
\NewDocumentCommand{\affineIndHat}{o}{
  \IfNoValueTF{#1}
    {\widehat{\mathcal{A}}}
    {\widehat{\mathcal{A}}^{(#1)}}}
\NewDocumentCommand{\affineDelta}{o}{
  \IfNoValueTF{#1}
    {\mathcal{A}^{\Delta}}
    {\mathcal{A}^{\Delta(#1)}}}

\NewDocumentCommand{\affineIndPrime}{o}{
  \IfNoValueTF{#1}
    {\mathcal{A}^{\prime}}
    {\mathcal{A}^{\prime(#1)}}}
\NewDocumentCommand{\affineIndHatPrime}{o}{
  \IfNoValueTF{#1}
    {\widehat{\mathcal{A}}^{\prime}}
    {\widehat{\mathcal{A}}^{\prime(#1)}}}

\newcommand{\reluIndLb}[2]{\mathcal{L}^{L(#1)}_{#2}}

\NewDocumentCommand{\reluIndLbz}{o}{
  \IfNoValueTF{#1}
    {\mathcal{L}^{L0}}
    {\mathcal{L}^{L0(#1)}}}
\NewDocumentCommand{\reluIndLbo}{o}{
  \IfNoValueTF{#1}
    {\mathcal{L}^{L1}}
    {\mathcal{L}^{L1(#1)}}}
\NewDocumentCommand{\reluIndUb}{o}{
  \IfNoValueTF{#1}
    {\mathcal{L}^{U}}
    {\mathcal{L}^{U(#1)}}}

\NewDocumentCommand{\reluIndHatLbz}{o}{
  \IfNoValueTF{#1}
    {\widehat{\mathcal{L}}^{L0}}
    {\widehat{\mathcal{L}}^{L0(#1)}}}
\NewDocumentCommand{\reluIndHatLbo}{o}{
  \IfNoValueTF{#1}
    {\widehat{\mathcal{L}}^{L1}}
    {\widehat{\mathcal{L}}^{L1(#1)}}}
\NewDocumentCommand{\reluIndHatUb}{o}{
  \IfNoValueTF{#1}
    {\widehat{\mathcal{L}}^{U}}
    {\widehat{\mathcal{L}}^{U(#1)}}}
\NewDocumentCommand{\reluRelUnsNeg}{o}{
  \IfNoValueTF{#1}
    {\mathcal{P}}
    {\mathcal{P}^{(#1)}}}
\newcommand{\reluRelUnsNegLb}[2]{\mathcal{P}^{L(#1)}_{#2}}

\NewDocumentCommand{\reluRelUnsNegLbz}{o}{
  \IfNoValueTF{#1}
    {\mathcal{P}^{L0}}
    {\mathcal{P}^{L0(#1)}}}
\NewDocumentCommand{\reluRelUnsNegLbo}{o}{
  \IfNoValueTF{#1}
    {\mathcal{P}^{L1}}
    {\mathcal{P}^{L1(#1)}}}
\NewDocumentCommand{\reluRelUnsNegUb}{o}{
  \IfNoValueTF{#1}
    {\mathcal{P}^{U}}
    {\mathcal{P}^{U(#1)}}}
\NewDocumentCommand{\reluRelNegUns}{o}{
  \IfNoValueTF{#1}
    {\mathcal{Q}}
    {\mathcal{Q}^{(#1)}}}

\NewDocumentCommand{\reluRelNegUnsLbz}{o}{
  \IfNoValueTF{#1}
    {\mathcal{Q}^{L0}}
    {\mathcal{Q}^{L0(#1)}}}
\NewDocumentCommand{\reluRelNegUnsLbo}{o}{
  \IfNoValueTF{#1}
    {\mathcal{Q}^{L1}}
    {\mathcal{Q}^{L1(#1)}}}
\NewDocumentCommand{\reluRelNegUnsUb}{o}{
  \IfNoValueTF{#1}
    {\mathcal{Q}^{U}}
    {\mathcal{Q}^{U(#1)}}}
\NewDocumentCommand{\reluRelUnsPos}{o}{
  \IfNoValueTF{#1}
    {\mathcal{R}}
    {\mathcal{R}^{(#1)}}}

\NewDocumentCommand{\reluRelUnsPosLbz}{o}{
  \IfNoValueTF{#1}
    {\mathcal{R}^{L0}}
    {\mathcal{R}^{L0(#1)}}}
\NewDocumentCommand{\reluRelUnsPosLbo}{o}{
  \IfNoValueTF{#1}
    {\mathcal{R}^{L1}}
    {\mathcal{R}^{L1(#1)}}}
\NewDocumentCommand{\reluRelUnsPosUb}{o}{
  \IfNoValueTF{#1}
    {\mathcal{R}^{U}}
    {\mathcal{R}^{U(#1)}}}
\NewDocumentCommand{\reluRelPosUns}{o}{
  \IfNoValueTF{#1}
    {\mathcal{S}}
    {\mathcal{S}^{(#1)}}}

\NewDocumentCommand{\reluRelPosUnsLbz}{o}{
  \IfNoValueTF{#1}
    {\mathcal{S}^{L0}}
    {\mathcal{S}^{L0(#1)}}}
\NewDocumentCommand{\reluRelPosUnsLbo}{o}{
  \IfNoValueTF{#1}
    {\mathcal{S}^{L1}}
    {\mathcal{S}^{L1(#1)}}}
\NewDocumentCommand{\reluRelPosUnsUb}{o}{
  \IfNoValueTF{#1}
    {\mathcal{S}^{U}}
    {\mathcal{S}^{U(#1)}}}
\NewDocumentCommand{\reluRelDeltaPos}{o}{
  \IfNoValueTF{#1}
    {\mathcal{T}}
    {\mathcal{T}^{(#1)}}}
\NewDocumentCommand{\reluRelDeltaPosLb}{o}{
  \IfNoValueTF{#1}
    {\mathcal{T}^{L}}
    {\mathcal{T}^{L(#1)}}}
\NewDocumentCommand{\reluRelDeltaPosUb}{o}{
  \IfNoValueTF{#1}
    {\mathcal{T}^{U}}
    {\mathcal{T}^{U(#1)}}}
\NewDocumentCommand{\reluRelDeltaNeg}{o}{
  \IfNoValueTF{#1}
    {\mathcal{V}}
    {\mathcal{V}^{(#1)}}}
\NewDocumentCommand{\reluRelDeltaNegLb}{o}{
  \IfNoValueTF{#1}
    {\mathcal{V}^{L}}
    {\mathcal{V}^{L(#1)}}}
\NewDocumentCommand{\reluRelDeltaNegUb}{o}{
  \IfNoValueTF{#1}
    {\mathcal{V}^{U}}
    {\mathcal{V}^{U(#1)}}}
\NewDocumentCommand{\reluRelDeltaUns}{o}{
  \IfNoValueTF{#1}
    {\mathcal{D}}
    {\mathcal{D}^{(#1)}}}
\NewDocumentCommand{\reluRelDeltaUnsLb}{o}{
  \IfNoValueTF{#1}
    {\mathcal{D}^{L}}
    {\mathcal{D}^{L(#1)}}}
\NewDocumentCommand{\reluRelDeltaUnsUb}{o}{
  \IfNoValueTF{#1}
    {\mathcal{D}^{U}}
    {\mathcal{D}^{U(#1)}}}

\newcommand{\objCoef}{\mu}
\newcommand{\constCoef}{\nu}
\newcommand{\constCoefDelta}{\lambda}
\newcommand{\objCoefInd}[2]{\objCoef^{(#1)}_{#2}}
\newcommand{\objCoefIndHat}[2]{\widehat{\objCoef}^{(#1)}_{#2}}

\newcommand{\objCoefDeltaPos}[2]{\objCoef^{+\Delta (#1)}_{#2}}
\newcommand{\objCoefDeltaNeg}[2]{\objCoef^{-\Delta (#1)}_{#2}}
\newcommand{\objCoefIndPrime}[2]{\objCoef^{\prime(#1)}_{#2}}
\newcommand{\objCoefIndHatPrime}[2]{\widehat{\objCoef}^{\prime(#1)}_{#2}}

\newcommand{\objCoefDeltaPosPrime}[2]{\objCoef^{\prime+\Delta (#1)}_{#2}}
\newcommand{\objCoefDeltaNegPrime}[2]{\objCoef^{\prime-\Delta (#1)}_{#2}}
\newcommand{\constCoefInd}[2]{\constCoef^{(#1)}_{#2}}
\newcommand{\constCoefIndDelta}[2]{\constCoef^{\Delta (#1)}_{#2}}
\newcommand{\constCoefIndHat}[2]{\widehat{\constCoef}^{(#1)}_{#2}}
\newcommand{\constCoefIndHatDelta}[2]{\widehat{\constCoef}^{\Delta (#1)}_{#2}}
\newcommand{\constCoefDeltaPos}[2]{\constCoefDelta^{+ (#1)}_{#2}}
\newcommand{\constCoefDeltaNeg}[2]{\constCoefDelta^{- (#1)}_{#2}}
\newcommand{\constCmplxInd}[2]{\pi^{(#1)}_{#2}}
\newcommand{\constCmplxIndHat}[2]{\widehat{\pi}^{(#1)}_{#2}}
\newcommand{\constCmplxDeltaL}[2]{\pi^{\Delta (#1)}_{#2 L}}
\newcommand{\constCmplxDeltaU}[2]{\pi^{\Delta (#1)}_{#2 U}}
\newcommand{\objCmplxInd}[2]{\omega^{(#1)}_{#2}}
\newcommand{\objCmplxIndHat}[2]{\widehat{\omega}^{(#1)}_{#2}}
\newcommand{\objCmplxIndDelta}[2]{\omega^{\Delta (#1)}_{#2}}

\NewDocumentCommand{\preReluInd}{o}{
  \IfNoValueTF{#1}
    {\mathcal{B}}
    {\mathcal{B}^{(#1)}}}
\NewDocumentCommand{\preReluIndHat}{o}{
  \IfNoValueTF{#1}
    {\widehat{\mathcal{B}}}
    {\widehat{\mathcal{B}}^{(#1)}}}
\NewDocumentCommand{\preReluDelta}{o}{
  \IfNoValueTF{#1}
    {\mathcal{B}^{\Delta}}
    {\mathcal{B}^{\Delta(#1)}}}

\NewDocumentCommand{\preCoefInd}{o}{
  \IfNoValueTF{#1}
    {c}
    {c^{(#1)}}}
\NewDocumentCommand{\preCoefIndHat}{o}{
  \IfNoValueTF{#1}
    {\widehat{c}}
    {\widehat{c}^{(#1)}}}
\NewDocumentCommand{\preCoefDeltaPos}{o}{
  \IfNoValueTF{#1}
    {c^{+\Delta}}
    {c^{+\Delta(#1)}}}
\NewDocumentCommand{\preCoefDeltaNeg}{o}{
  \IfNoValueTF{#1}
    {c^{-\Delta}}
    {c^{-\Delta(#1)}}}

\NewDocumentCommand{\preCoefIndPrime}{o}{
  \IfNoValueTF{#1}
    {c^{\prime}}
    {c^{\prime(#1)}}}
\NewDocumentCommand{\preCoefIndHatPrime}{o}{
  \IfNoValueTF{#1}
    {\widehat{c}^{\prime}}
    {\widehat{c}^{\prime(#1)}}}
\NewDocumentCommand{\preCoefDeltaPosPrime}{o}{
  \IfNoValueTF{#1}
    {c^{\prime+\Delta}}
    {c^{\prime+\Delta(#1)}}}
\NewDocumentCommand{\preCoefDeltaNegPrime}{o}{
  \IfNoValueTF{#1}
    {c^{\prime-\Delta}}
    {c^{\prime-\Delta(#1)}}}

\NewDocumentCommand{\coefInd}{o}{
  \IfNoValueTF{#1}
    {\nu}
    {\nu^{(#1)}}}
\NewDocumentCommand{\coefIndHat}{o}{
  \IfNoValueTF{#1}
    {\widehat{\nu}}
    {\widehat{\nu}^{(#1)}}}
\NewDocumentCommand{\coefDeltaPos}{o}{
  \IfNoValueTF{#1}
    {\nu^{+\Delta}}
    {\nu^{+\Delta(#1)}}}
\NewDocumentCommand{\coefDeltaNeg}{o}{
  \IfNoValueTF{#1}
    {\nu^{-\Delta}}
    {\nu^{-\Delta(#1)}}}

\NewDocumentCommand{\dualIndReluLbZ}{o}{
  \IfNoValueTF{#1}
    {\mu}
    {\mu^{(#1)}}}
\NewDocumentCommand{\dualIndReluLbO}{o}{
  \IfNoValueTF{#1}
    {\tau}
    {\tau^{(#1)}}}
\NewDocumentCommand{\dualIndReluUb}{o}{
  \IfNoValueTF{#1}
    {\lambda}
    {\lambda^{(#1)}}}

\newcommand{\network}{\ensuremath{N}\xspace} 

\newcommand{\R}{\ensuremath{\mathbb{R}}\xspace}
\newcommand{\outLayerIdx}{L} 
\newcommand{\inpSpace}{\ensuremath{\Omega}\xspace}
\newcommand{\inpDim}{\ensuremath{n}\xspace}
\newcommand{\inpShape}{\ensuremath{\R^{\inpDim}}\xspace} 
\newcommand{\outDim}{\ensuremath{m}\xspace}
\newcommand{\outShape}{\R^{\outDim}} 
\newcommand{\TRUE}{\ensuremath{\mathsf{true}}\xspace}
\newcommand{\FALSE}{\ensuremath{\mathsf{false}}\xspace}
\newcommand{\queue}{\ensuremath{Q}\xspace}
\newcommand{\transposed}{{}^T}
\newcommand{\uid}[1]{^{(#1)}} 
\newcommand{\weight}[2]{\ensuremath{W^{(#1)}_{#2}}\xspace}
\newcommand{\biase}[2]{\ensuremath{B^{(#1)}_{#2}}\xspace}
\newcommand{\relu}{ReLU\xspace}
\newcommand{\relus}{ReLUs\xspace}
\newcommand{\inp}{\ensuremath{y}\xspace}
\newcommand{\inpOne}{\ensuremath{\inp}\xspace} 
\newcommand{\inpTwo}{\ensuremath{\hat{\inp}}\xspace} 
\newcommand{\PostIndNeuron}{\ensuremath{y}\xspace} 
\newcommand{\PreIndNeuron}{\ensuremath{x}\xspace} 
\newcommand{\indNeuron}[2]{\ensuremath{\PostIndNeuron^{(#1)}_{#2}}\xspace}
\newcommand{\indNeuronPre}[2]{\ensuremath{\PreIndNeuron^{(#1)}_{#2}}\xspace} 
\newcommand{\indNeuronTwoPost}[2]{\ensuremath{\hat{\PostIndNeuron}^{(#1)}_{#2}}\xspace}
\newcommand{\indNeuronTwoPre}[2]{\ensuremath{\hat{\PreIndNeuron}^{(#1)}_{#2}}\xspace}
\newcommand{\prtbInpOne}{\ensuremath{\inp^{\prime}}\xspace} 
\newcommand{\indNeuronOne}[2]{\ensuremath{y^{(#1)}_{#2}}\xspace}  
\newcommand{\indNeuronTwo}[2]{\ensuremath{\hat{y}^{(#1)}_{#2}}\xspace}  
\newcommand{\relNeuron}{\Delta} 
\newcommand{\relNeuronPre}[2]{\relNeuron x^{(#1)}_{#2}}
\newcommand{\relNeuronPost}[2]{\relNeuron y^{(#1)}_{#2}}
\newcommand{\tool}{\textsc{SaBRe}\xspace}
\newcommand{\bab}{\textsf{BaB}\xspace}
\newcommand{\raven}{\textsf{RaVeN}\xspace} 
\newcommand{\diffpoly}{\ensuremath{\mathtt{DiffPoly}}\xspace}

\newcommand{\rabbit}{\textsf{RABBit}\xspace} 
\newcommand{\babsr}{\textsf{BaBSR}\xspace} 
\newcommand{\appVerifierInd}{\ensuremath{\mathsf{ApxVer}}\xspace}
\newcommand{\appVerifierRel}{\ensuremath{\mathsf{ApxVerRel}}\xspace}

\newcommand{\cex}[1]{\ensuremath{\widehat{#1}}\xspace}
\newcommand{\inpSpecDist}{\ensuremath{\varepsilon}\xspace} 
\newcommand{\outSpecDist}{\ensuremath{\delta}\xspace} 
\newcommand{\relspec}{\ensuremath{\Phi}\xspace}
\newcommand{\baseline}{\textsf{ClasIS}\xspace}
\newcommand{\is}{\textsf{DualIS}\xspace}
\newcommand{\RSrandom}{\textsf{RandRS}\xspace}
\newcommand{\deltaTime}{\ensuremath{\Delta \textsc{t}}\xspace}
\newcommand{\subproblems}{\ensuremath{\texttt{p}^\#}\xspace}
\newcommand{\solved}{\ensuremath{\texttt{s}^\#}\xspace}
\newcommand{\perturbRate}{\ensuremath{\texttt{p\%}}\xspace}
\newcommand{\mnist}{\textsf{MNIST}\xspace}
\newcommand{\mnistF}{\textsf{MNIST-F}\xspace}
\newcommand{\mnistC}{\textsf{MNIST-C}\xspace}
\newcommand{\acasxu}{\textsf{ACAS Xu}\xspace}
\newcommand{\cifar}{\textsf{CIFAR}\xspace}

\newcommand{\cifarTen}{\textsf{CIFAR-10}\xspace}

\newcommand{\gtsrb}{\textsf{GTSRB}\xspace}
\newcommand{\lowerBound}[1]{\ensuremath{\underline{#1}}\xspace}
\newcommand{\upperBound}[1]{\ensuremath{\overline{#1}}\xspace}
\newcommand{\inpSpaceUpper}[1]{\ensuremath{\upperBound{\inpSpace_{#1}}}\xspace}
\newcommand{\inpSpaceLower}[1]{\ensuremath{\lowerBound{\inpSpace_{#1}}}\xspace}
\newcommand{\posRel}[1]{\ensuremath{\left(#1\right)^{+}}\xspace}
\newcommand{\negRel}[1]{\ensuremath{\left(#1\right)^{-}}\xspace}
\newcommand{\labelIdx}{\ensuremath{\ell}\xspace}
\newcommand{\relConLen}{\ensuremath{k}\xspace}
\newcommand{\slope}[2]{\alpha^{(#1)}_{#2}}
\newcommand{\slopeHat}[2]{\hat{\alpha}^{(#1)}_{#2}}
\newcommand{\exampleCase}{\ensuremath{j \in \reluUns_i \cap \reluNegHat_i}\xspace}
\newcommand{\lagrFunc}{\ensuremath{LF}\xspace}
\newcommand{\kkt}{KKT\xspace}
\newcommand{\dualVar}{\ensuremath{\theta}\xspace}
\newcommand{\equC}{\ensuremath{C_{equ}}\xspace}
\newcommand{\ineC}{\ensuremath{C_{ine}}\xspace}
\newcommand{\constE}{\ensuremath{ce}\xspace}  
\newcommand{\constI}{\ensuremath{ci}\xspace}  
 \newcommand{\constInput}{\ensuremath{\mathtt{Const_{inp}}}\xspace}
\newcommand{\constAffine}{\ensuremath{\mathtt{Const_{affn}}}\xspace}
\newcommand{\constRelAffine}{\ensuremath{\mathtt{Const_{\Delta affn}}}\xspace}
\newcommand{\constIndRelu}{\ensuremath{\mathtt{Const_{ReLU}}}\xspace}
\newcommand{\constRelRelu}{\ensuremath{\mathtt{Const_{\Delta ReLU}}}\xspace}

\newcommand\mathcompact
{%
\thinmuskip= 0mu                  
\medmuskip= 0mu 
\thickmuskip= 0mu        
}

\newcommand{\myparagraph}[1]{\medskip\noindent{\bf #1.}}

\usepackage[hidelinks]{hyperref}

\begin{document}

\title{Branch and Bound for Relational Verification of Neural Networks}

\author{Kota Fukuda, Zhenya Zhang, Guanqin Zhang and Jianjun Zhao
\thanks{K. Fukuda, Z. Zhang and J. Zhao are with Graduate School and Faculty of Information Science and Electrical Engineering, Kyushu University, Fukuoka, Japan. Z. Zhang is also with National Institute of Informatics, Tokyo, Japan. E-mail: fukuda.kota.527@s.kyushu-u.ac.jp, \{zhang, zhao\}@ait.kyushu-u.ac.jp}
\thanks{G. Zhang is with UNSW Sydney, Australia. E-mail: guanqin.zhang@unsw.edu.au}
\thanks{Manuscript received March 30, 2026; revised June 19, 2026; accepted July 17, 2026.}
}




\maketitle

\begin{abstract}
Verification of neural networks against relational specifications, such as global robustness, is crucial for safety-critical applications of cyber-physical systems (CPS), given their increasing adoption of AI components. 
Compared to simple trace properties (e.g., local robustness), verifying relational specifications requires reasoning about the relationship between multiple network inferences, which brings significant technical challenges. Existing research has explored abstraction techniques based on sound and convex over-approximation of neural network outputs; however, since these approaches are inherently incomplete and may raise false alarms, they further underscore the need of effective abstraction refinement.

In this paper, we propose a \emph{branch-and-bound (\bab)} framework to mitigate the issue, which iteratively splits the problem until all sub-problems are verified. Specifically, our \bab framework features splitting of relational neurons rather than individual neurons as prior works do, and as the core of our technique, we devise a relational neuron selection strategy based on the dual formulation of the verification problem, which allows us to efficiently select the (most likely) optimal relational neuron that maximizes the refinement brought by problem splitting.
We evaluate \tool on 817 verification problems across \acasxu, \mnistF, \mnistC, \cifar and \gtsrb. The results show that \tool outperforms different baseline approaches, in terms of the number of solved instances and verification efficiency, which demonstrates the effectiveness of our proposed techniques.
\end{abstract}


\section{Introduction}\label{sec:introduction}
Neural networks have been increasingly adopted in safety-critical domains such as autonomous driving, healthcare, power grids, and robotics. Owing to their ability to learn highly nonlinear functions, they have achieved remarkable performance gains over traditional model-based approaches and have emerged as promising solutions across various components of cyber-physical systems (CPS) including the central control units such as perception and planning. Meanwhile, due to the safety-critical nature of CPS, significant concerns arise about the unpredictable behavior of neural networks. For example, existing research has shown that they can easily make wrong predictions under small input perturbations~\cite{biggio2013evasion, goodfellow2015explaining}. As central control units in CPS, reliability of neural network components is of utmost importance, and therefore formal verification is a preferred approach capable of certifying their robustness against unforeseen perturbations.

Most existing works~\cite{katz2017reluplex,huang2017safety,henriksen2021deepsplit,singh2019abstract,bunel2020branch, wang2021beta, singh2018deepz} in neural network verification take local robustness as their target, which requires the neural network to predict any perturbed input correctly within an $\ell_p$-ball of a clean input. However, local robustness has limited expressivity; for example, a neural network controller for an adaptive cruise control system~\cite{zhang2022falsifai} is expected to be \emph{globally robust}, i.e., given similar environmental inputs, its control actions should not change drastically. It is worth noting that, global robustness is a particularly relevant, and often more demanding property, in the context of CPS, because control systems often require behavioral consistency at a broad scale, e.g., a wide region of the input space. These properties give rise to the \emph{relational verification} problem, which requires reasoning about the relationships between multiple network inferences and thus cannot be directly handled by techniques designed for local robustness verification. 

Despite its importance, relational verification remains much underexplored, compared to classic verification against local robustness. Nevertheless, standing on the shoulder of giants, existing research~\cite{wang22itne, banerjee24raven} has explored approximation approaches, which aim to derive a convex abstraction domain to soundly over-approximate the  output bounds, in order to cope with the non-linearity of neural network inferences. In particular, instead of each individual inference, these approaches aim to approximate the \emph{relationship} between different inferences, which can be characterized by \emph{relational neurons}, each symbolizing the difference between the outputs of the same neuron over different inferences.  

While approximation approaches are sound and efficient, they suffer from a completeness issue, i.e., they may raise false alarms, reporting violations that do not actually exist. \emph{Branch and bound (\bab)}~\cite{bunel2020branch} is a widely-recognized approach for abstraction refinement, and has been extensively explored~\cite{henriksen2021deepsplit,de2021improved,shi2024neural,luneural,ferrari2022complete,wang2021beta} in classic neural network verification. In the context of relational verification, it has been studied in~\cite{suresh24rabbit} to handle \emph{universal adversarial perturbations}, together with some domain-specific heuristics dedicated to the specification. However, due to the problem setting, their approach focuses on individual inferences rather than relationship between inferences, so they may not achieve optimal performances.

\myparagraph{Contributions} 
In this paper, we propose a novel \bab-based verification framework \tool, standing for \underline{\textbf{S}}plitting \underline{\textbf{A}}pproximated \underline{\textbf{B}}ounds for \underline{\textbf{RE}}lational verification. In line with classic \bab, it also aims to reduce approximation error by iterative problem splitting; differently, due to the specific problem setting of relational verification, \tool targets relational neurons rather than individual neurons for problem splitting. Consequently, our approach remains approximative and is therefore incomplete. Nevertheless, because relational neurons have significant impact on the precision of approximated output bounds, splitting them can maximize the degree of refinement and thereby improve verification efficiency.

There remains a critical factor that affects the efficiency of \bab, namely, the choice of which neuron to split at each step of \bab. To address this, we design a relational neuron selection strategy based on estimating the improvement in verification precision induced by splitting different neurons. Specifically, following the approach~\cite{bunel2020branch}, we formulate the Lagrangian dual problem in the relational setting, whose objective function soundly approximates the output difference bound. This formulation enables efficient estimation of the impact of splitting each relational neuron, allowing us to select the one that is most likely to maximize bound refinement.

We perform experiments to evaluate our proposed approach, on 817 verification problem instances from commonly-adopted datasets, including \acasxu, \mnist, \cifarTen, and \gtsrb. The results demonstrate that our proposed framework significantly outperforms existing \bab approaches that rely on splitting of individual neurons, in both verification time and the number of solved instances. Moreover, our study also shows the effectiveness of the proposed neuron selection via a comparison with a random baseline. These results demonstrate the strengths of our approach, marking an important step in relational verification.

\section{Preliminaries}\label{sec:preliminaries}
We state our problem setting of relational verification with a focus on \emph{global robustness} properties, and then we introduce a recent solution~\cite{banerjee24raven} that proposes to use an approximation technique to solve the problem.

\subsection{Relational Verification Problem}\label{sec:problemStatement}
We follow existing literature~\cite{wang22itne, banerjee24raven} and take \emph{feed-forward neural networks} as our target systems. 
\begin{definition}[Neural networks]
    A \emph{(feed-forward) neural network} $\network: \inpShape \to \outShape$ maps an $\inpDim$-dim vector to an $\outDim$-dim vector, by $\outLayerIdx$ hidden layers. At the $i$-th hidden layer, it maps  $\indNeuron{i-1}{}\in\R^{n_{i-1}}$ to $\indNeuron{i}{}\in\R^{n_i}$ (where $n_0 = n$, $n_{\outLayerIdx} = \outDim$) by alternating between \emph{affine} and \emph{activation functions}, as follows, 
    \begin{align*}
        \indNeuronPre{i}{} = \weight{i-1}{}\indNeuron{i-1}{}+\biase{i}{}, \quad \indNeuron{i}{} = \sigma(\indNeuronPre{i}{})
    \end{align*}
    where $\weight{i-1}{}\in\R^{n_{i}\times n_{i-1}}$ and $\biase{i}{}\in\R^{n_i}$ are parameters at $i$-th layer and $\sigma$ is a non-linear function, such as ReLU: $\sigma(x) = \max(0, x)$. We denote by $\indNeuron{i}{j}\in\R$ the $j$-th component of $\indNeuron{i}{}$, which is the output of the $j$-th \emph{neuron} at $i$-th layer.
\end{definition}

\medskip
Most existing verification works~\cite{liu2021algorithms,yang2021improving,anderson2019optimization,zhao2022cleverest} consider \emph{local robustness} as the specification to evaluate, especially in classification tasks, which can be described as follows.

\begin{definition}[Local robustness]\label{def:localRob}
    Given $\network: \inpShape \to \outShape$ a neural network,
$\inpOne \in \inpShape$ a reference input, and $\inpSpecDist>0$ a perturbation
radius.
We define that $\network$ is \emph{$\inpSpecDist$-robust at $\inpOne$} if the following condition holds:
\begin{align*}\label{eq:def-loc-robust}
  & \textstyle \forall\,\prtbInpOne\in \inpShape,\;
   \|\,\prtbInpOne-\inpOne\,\|_{p} \le \inpSpecDist \\
   & \;\Longrightarrow\;
   \arg\max_{1\le j\le m} \network_{j}(\prtbInpOne) = \arg\max_{1\le j\le m} \network_{j}(\inpOne),
\end{align*}
where $\|\cdot\|_{p}$ a chosen norm, e.g., $p = \infty$, 
and $\network_{j}(\inpOne)$ denotes the $j$-th component of $\network(\inpOne)$. In other words, any perturbed input $\prtbInpOne$ from the reference input $\inpOne$ such that $\|\,\prtbInpOne-\inpOne\,\|_{p}\le\epsilon$
should be classified to the same label as that of $\inpOne$.
\end{definition}

\medskip
Essentially, Def.~\ref{def:localRob} is a \emph{safety property} that concerns with the reachability of the neural network output. However, it cannot describe more complex \emph{relational specifications}, such as \emph{global robustness}, that reason about the relationship between multiple neural network inferences. In this paper, we target global robustness as a representative relational specification to showcase our approach, based on the definition in~\cite{wang22itne}.

\begin{definition}[Global robustness~\cite{wang22itne}]\label{def:glb_rbst}
Given a neural network $\network : \inpShape \to \outShape$ and parameters $(\inpSpecDist, \outSpecDist)$, let $\inpSpace = \prod_{1\le j\le \inpDim}[\inpSpaceLower{j}, \inpSpaceUpper{j}]$ be an $\inpDim$-dim rectangular input space, where $\inpSpaceLower{j}$ and $\inpSpaceUpper{j}$ are respectively the lower and upper bounds in each dimension $j$.

$\network$ is  said to be
\textit{$(\inpSpecDist, \outSpecDist)$-globally robust}  w.r.t. the $\labelIdx$-th dimension,
if the following condition holds:
\begin{equation}
\forall \inpOne, \inpTwo \in \inpSpace,\quad
\|\inpOne - \inpTwo\|_\infty \le \inpSpecDist
\implies
\left|\network_{\labelIdx}(\inpOne) - \network_{\labelIdx}(\inpTwo)\right| \le \outSpecDist.
\end{equation}
\end{definition}

The essential difference of global robustness in Def.~\ref{def:glb_rbst} from local robustness in Def.~\ref{def:localRob} consists in that, while in Def.~\ref{def:localRob},  $\prtbInpOne$ is the only variable to deal with, in Def.~\ref{def:glb_rbst}, both $\inpOne$ and $\inpTwo$ are variables necessary to deal with. These specifications are known as a type of \emph{hyperproperties}~\cite{clarkson2010hyperproperties}. In this paper, we aim to solve the following verification problem against global robustness.

\begin{definition}[Global robustness verification problem] \label{def:problemStatement}
Given the above definitions, 
the global robustness verification problem is to determine whether  a neural network
$\network$ satisfies the $(\inpSpecDist, \outSpecDist)$-global robustness 
condition in Def.~\ref{def:glb_rbst}.
\end{definition}

\subsection{Verification by Over-Approximation}
\label{sec:OverAppVerification}
Unlike local robustness verification, relational verification requires reasoning about multiple inferences, so most existing approaches~\cite{wang22itne, banerjee24raven, xie2023deepgemini} solve the problem by producing multiple copies of the network, each processing one inference. Then, given the setting in Def.~\ref{def:glb_rbst}, the problem can be solved by maximizing the difference of network output---if the maximum of the difference is less than $\outSpecDist$, we can verify that the network $\network$ satisfies global robustness. 

To solve this problem efficiently, existing research~\cite{wang22itne,banerjee24raven} explores the use of approximation strategies that can soundly over-approximate the differences between network outputs for different inferences, thereby verifying the problem. 

\myparagraph{Over-approximation approach}
Classic verification against local robustness considers a single network inference, and in that case, 
an approximation verifier can be considered as a function $\appVerifierInd$ that, given a neural network $\network$ and an input space $\inpSpace$, returns a range $\left[\lowerBound{\indNeuron{i}{j}}, \upperBound{\indNeuron{i}{j}}\right]$ for each neuron $\indNeuron{i}{j}$ that over-approximates the possibly reachable range of $\indNeuron{i}{j}$ under any input $\indNeuron{0}{}\in \inpSpace$ of the network.
 In contrast, relational verification needs to reason about multiple inferences. Below, we denote by $\indNeuronOne{i}{j}$ and $\indNeuronTwo{i}{j}$ respectively the neuron output of each individual inference, and $\relNeuronPost{i}{j}$ the difference between the $\indNeuronOne{i}{j}$ and $\indNeuronTwo{i}{j}$, i.e., $\relNeuronPost{i}{j} = \indNeuronOne{i}{j} - \indNeuronTwo{i}{j}$. Notably, in addition to the original neurons $\indNeuronOne{i}{j}$ and $\indNeuronTwo{i}{j}$  (which we call \emph{individual neurons}),  the differences $\relNeuronPost{i}{j}$ between corresponding individual neurons constitute additional symbolic nodes, which we call \emph{relational neurons}. Fig.~\ref{fig:apprx_propagation} visualizes these two types of neurons, in which each circle represents an individual neuron, and each square represents a relational neuron.

Accordingly, an approximation verifier for relational verification can be deemed as a function \appVerifierRel that, given a neural network $\network$, an input space \inpSpace, and a real parameter $\inpSpecDist$, returns  not only the ranges for individual neurons $\indNeuronOne{i}{j}$ and $\indNeuronTwo{i}{j}$, but also a range $\left[\lowerBound{\relNeuronPost{i}{j}}, \upperBound{\relNeuronPost{i}{j}} \right]$ for each relational neuron $\relNeuronPost{i}{j}$; similarly, this range  over-approximates the possibly reachable region of each relational neuron, under any pair of inputs $\indNeuronOne{0}{}, \indNeuronTwo{0}{}\in \inpSpace$ that hold $\|\indNeuronOne{0}{} - \indNeuronTwo{0}{}\|_\infty \le \inpSpecDist$.

\begin{wrapfigure}[12]{r}{0.36\linewidth}
\centering\includegraphics[width=\linewidth]{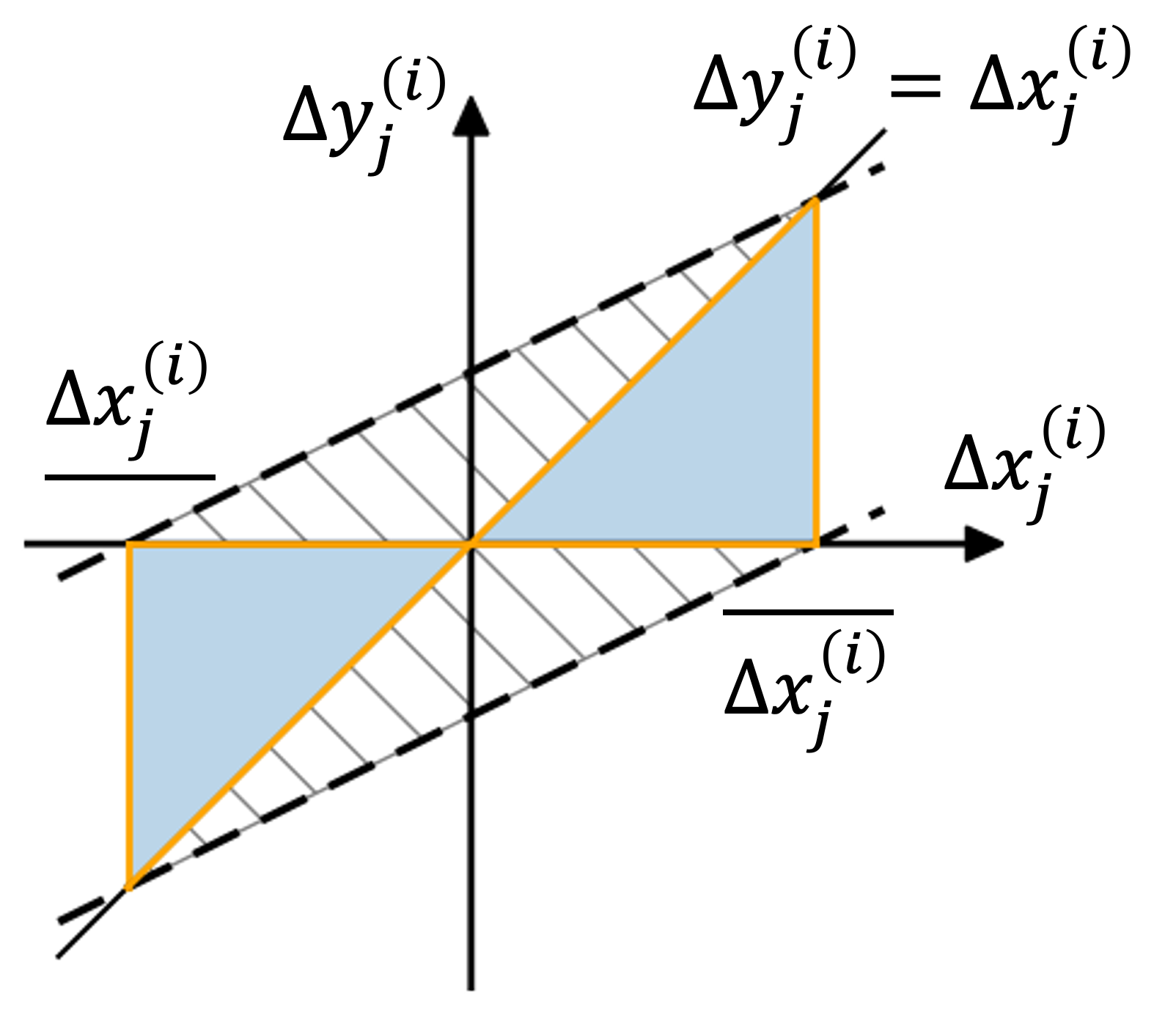}
    \caption{Linear relaxation of the difference of ReLU's outputs}
    \label{fig:diffpoly}
\end{wrapfigure}

Below, we elaborate on a recent approach \raven~\cite{banerjee24raven} to showcase how  \appVerifierRel works. In particular, \raven adopts a convex abstraction domain for approximating the bounds of relational neurons. By the domain, it can propagate the bounds of each relational neuron layer by layer, from the input to the output. The propagation process is briefed as follows:

\begin{compactitem}[$\bullet$]
    \item At input layer, the difference $\relNeuronPost{0}{}$ is bounded by $\inpSpecDist$, i.e., $\relNeuronPost{0}{} \in [-\inpSpecDist, \inpSpecDist]^{\inpDim}$;
    \item At the $i$-th layer, 
    \begin{compactenum}[i)]
        \item Given the difference $\relNeuronPost{i-1}{}$ of the previous layer, $\relNeuronPre{i}{}$ is obtained by the affine function $\relNeuronPre{i}{} = \weight{i-1}{}\relNeuronPost{i-1}{}$, so the bounds of $\relNeuronPre{i}{}$ can also be obtained from the bounds of $\relNeuronPost{i-1}{}$ accordingly by applying the affine transformation;
        \item Then, given $\relNeuronPre{i}{}$, since $\relNeuronPost{i}{}$ is obtained by a non-linear transformation,  the bounds of $\relNeuronPost{i}{}$ need to be computed by linear relaxation. To this end, \raven adopts a strategy used in~\cite{wang22itne}, which is visualized in Fig.~\ref{fig:diffpoly} and formulated as follows. Let $\relNeuronPre{i}{j} \in \left[\lowerBound{\relNeuronPre{i}{j}}, \upperBound{\relNeuronPre{i}{j}}\right]$, where $\lowerBound{\relNeuronPre{i}{j}}$ and $\upperBound{\relNeuronPre{i}{j}}$ are respectively the lower and upper bounds of $\relNeuronPre{i}{j}$; then, $\relNeuronPost{i}{j}$ is bounded by the following two linear constraints, i.e., the two dashed lines in Fig.~\ref{fig:diffpoly}.
\begin{equation}\label{eq:raven_rel_apprx}
        \begin{aligned}
            &\relNeuronPost{i}{j} \in \\
            &\left[\frac{\lowerBound{\relNeuronPre{i}{j}}(\upperBound{\relNeuronPre{i}{j}} - \relNeuronPre{i}{j})}{\upperBound{\relNeuronPre{i}{j}} - \lowerBound{\relNeuronPre{i}{j}}}, \;\; \frac{\upperBound{\relNeuronPre{i}{j}}(\relNeuronPre{i}{j} - \lowerBound{\relNeuronPre{i}{j}})}{\upperBound{\relNeuronPre{i}{j}} - \lowerBound{\relNeuronPre{i}{j}}}\right]
        \end{aligned}
        \end{equation}
    \end{compactenum}
\end{compactitem}

Layer by layer, we can finally obtain the bounds of the relational neurons $\relNeuronPost{\outLayerIdx}{}$ of the output layer, by which we can decide whether the network satisfies the given specification. 

In \raven, there is actually more heuristics that takes into account the bounds of individual neurons to further tighten the bounds of relational neurons. For instance, if the ReLU inputs $\indNeuronPre{i}{j}$ and $\indNeuronTwoPre{i}{j}$ of two corresponding individual neurons are both negative, despite their difference $\relNeuronPre{i}{j}$, the ReLU output difference $\relNeuronPost{i}{j}$ must be 0 because both $\indNeuronOne{i}{j}$ and $\indNeuronTwo{i}{j}$ are 0. 
As these details are not our main focus, we skip them and refer interested readers to~\cite{banerjee24raven}.

 Notably, if \appVerifierRel decides that a network is not globally robust, it will return a counterexample, i.e., a pair of inputs $\left(\cex{\indNeuronOne{0}{}}, \cex{\indNeuronTwo{0}{}}\right)$, as a witness of the violation to the specification. However, this counterexample may be a spurious one (often called a \emph{false alarm}) that actually does not violate the specification, due to the over-approximation of \appVerifierRel. 
To validate whether a counterexample is real, we can feed it to the neural network \network and check whether their outputs violate the requirement of global robustness. 

\section{Motivations}\label{sec:motivations}
We first review \emph{\bab}~\cite{bunel2020branch}, a state-of-the-art approach in classic neural network verification to refine the approximation brought by approximation approaches, such as \raven in~\S{}\ref{sec:OverAppVerification}, and then we use an example to motivate the advantages of considering relational neurons, rather than individual neurons as existing approaches do, when applying \bab to solve relational verification problems. 

\subsection{\bab for Abstraction Refinement}\label{sec:bab}
While approximation verifiers \appVerifierRel, such as \raven in~\S{}\ref{sec:OverAppVerification}, can provide sound over-approximation and are computationally efficient, they suffer from the \emph{completeness issue}: as mentioned, the approximated bounds contain the region not reachable by the network, so \appVerifierRel may raise spurious counterexamples (i.e., false alarms) that report a violation actually not existent.
To mitigate this issue, we need to refine the approximation in order to obtain more precise bounds, and to date,  \emph{\bab}~\cite{bunel2020branch} is the state-of-the-art approach in classic neural network verification against local robustness to address this issue. Below, we briefly review the workflow of \bab in the classic context.

\myparagraph{Branch and Bound (\bab)}\label{sec:bab_local}
Essentially, \bab involves a \emph{``divide-and-conquer''} strategy, which refines approximation by iterative problem splitting and application of verifiers to sub-problems. 
As approximation verifiers often introduce less approximation error for sub-problems, \bab can effectively tighten the approximation bounds and improve the precision of verification results. 
For local robustness, problem splitting is often performed with respect to the ReLU of a selected neuron, namely, it considers separately the cases when the input is positive and when the input is negative for the selected ReLU; by doing so, it can refine the approximation for the selected ReLU, because for each of the cases, ReLU becomes linear so there is no need to approximate its output. 

We review the workflow of \bab in classic verification against local robustness. 
\begin{compactenum}[i)]
    \item \label{step:check} Initially, \bab applies the verifier to the original problem, and checks whether the problem is verified: if yes, it terminates the verification and returns \TRUE;  
    \item \label{step:split} In case it raises a false alarm, \bab splits the problem into two sub-problems. Given ReLU as activation function, as mentioned, this is often achieved by imposing additional constraints to the input of a selected ReLU such that the transformation becomes linear.  
    \item For each of the sub-problems, \bab iteratively goes through Step~\ref{step:check} and Step~\ref{step:split} until all the sub-problems are verified, in which case it returns $\TRUE$. It is also possible that \bab encounters a real counterexample during the verification process, in which case it returns \FALSE.
\end{compactenum}
Consequently, \bab exploits the strength of approximation verifiers in efficiency and meanwhile complements their completeness issues to strike a balance.

\myparagraph{Neuron selection}\label{sec:neuron_selection_babsr}
During the \bab process, a key factor determining its efficiency is the selection of neurons to split at each step. Since neurons have different positions in the network architecture and different input ranges, splitting different neurons can bring different impacts on the final output bounds. To this end, neuron selection aims to maximize such impact and thereby to tighten the output bounds, such that it can verify the problem more efficiently.

Direct estimation of such impact consists in solving verification of different sub-problem candidates, which is not feasible. To address this issue, existing approaches, such as \babsr~\cite{bunel2020branch}, propose to utilize the Lagrangian dual formulation~\cite{wong2018provable} for estimation and are known to deliver strong performance.
In this approach, a primal problem consists in a linear program that includes all linear constraints for verifying a neural network (based on some specific linear relaxation for ReLU, e.g., by~\cite{singh2019abstract}). 
Then, in the dual formulation, the objective function can soundly over-approximate the output bounds, by a summation of neuron-wise terms, where each term captures the contribution of a specific neuron to the final output under the adopted linear relaxation. 
These terms are naturally collected during the construction of the dual formulation once the over-approximated bounds for each neuron are already available. 

Therefore, by choosing the terms associated with a candidate neuron to split, we can estimate the splitting effect without introducing additional overhead.

\subsection{Splitting of Relational Neurons}\label{sec:splitting_of_relational_neurons}
While \bab has been extensively studied in verification against local robustness, it remains much underexplored in relational verification. As an early work, \rabbit~\cite{suresh24rabbit} adopts \bab for verification against \emph{universal adversarial perturbations}, which is a different type of hyperproperties from the global robustness considered by us. Similarly to classic \bab, \rabbit also selects an individual neuron to split in case approximation verifiers raise a false alarm, and then deal with sub-problems until all of them are verified. 

In relational verification, in addition to individual neurons, relational neurons may also be selected to split. For example, for the abstraction domain in Fig.~\ref{fig:diffpoly}, problem splitting means to  consider the cases when $\relNeuronPre{i}{j} < 0$ and when $\relNeuronPre{i}{j} > 0$ separately; as each of the cases involves a reachable region as a blue triangle, the orange lines are sufficient to bound each of them. As a consequence, we can eliminate the shadowed region brought by the linear relaxation in \raven and thereby tighten the relational bounds. Existing work, such as \rabbit, has shown the effectiveness of \bab in relational verification, however, as it targets \emph{individual neurons} rather than \emph{relational neurons}, it fails to exploit the problem structure of relational verification in which the bounds of relational neurons are relevant to the final outputs and thus may not be able to achieve the optimal performance.

Below, we use an example to demonstrate the strength of splitting relational neurons.  The code of the example can be found in our online repository~\cite{SupplementaryMaterialEMSOFT2026}.

\begin{figure}[!tb]
    \begin{subfigure}{\linewidth}
    \centering\includegraphics[width=\linewidth]{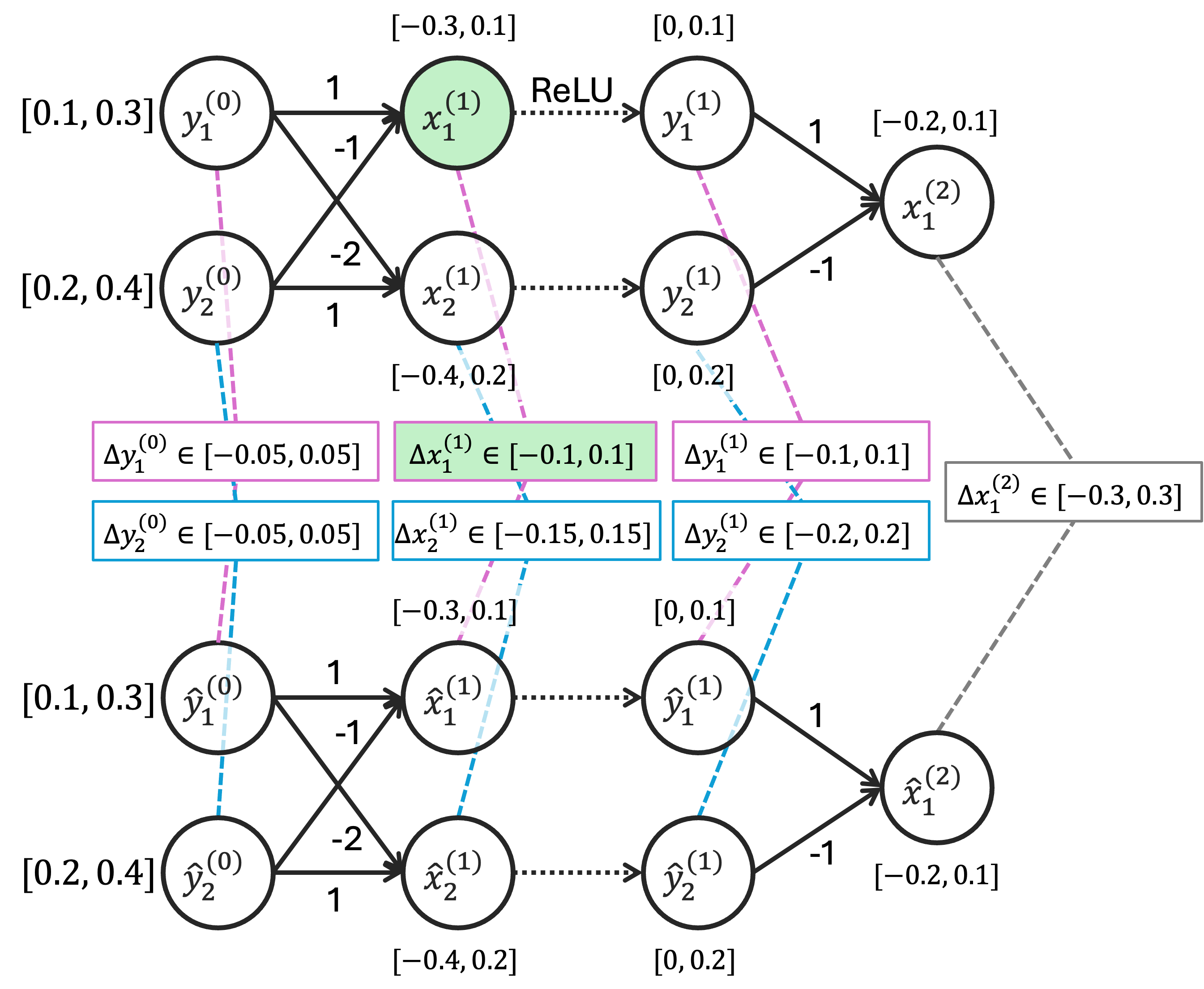}
    \caption{Approximate symbolic bounds propagation}
    \label{fig:apprx_propagation}
     \end{subfigure}
 \begin{subfigure}{\linewidth}
\centering\includegraphics[width=1.0\linewidth]{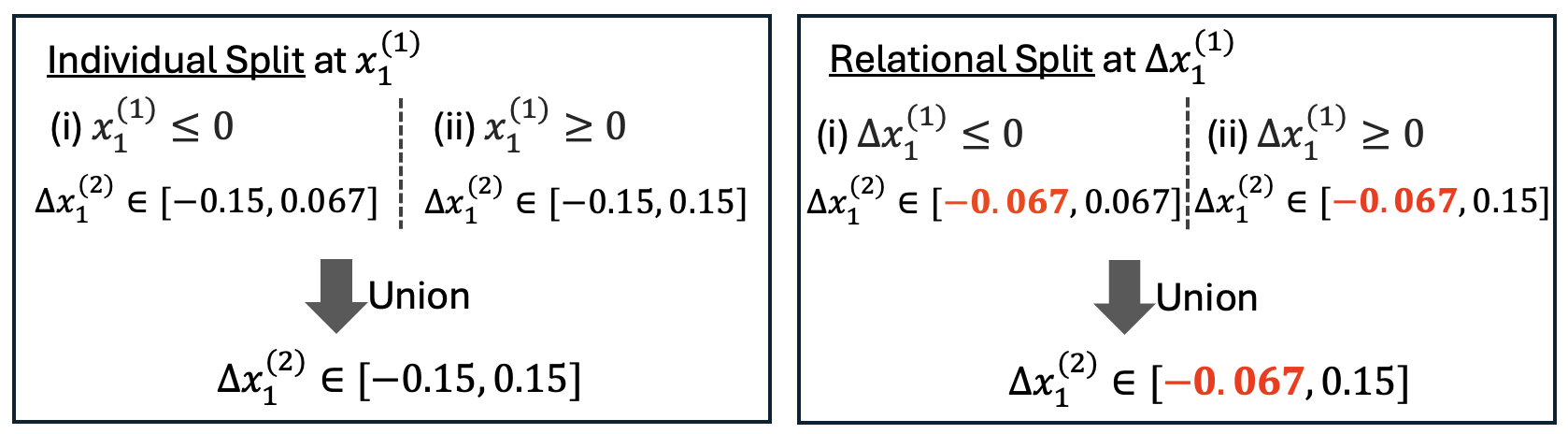}
    \caption{Splitting of individual neurons vs. relational neurons (solved by Gurobi~\cite{gurobi} following our approach in~\S{}\ref{sec:approach})}
    \label{fig:after_split}
 \end{subfigure}
 \caption{Symbolic bounds propagation and comparison between splitting of individual and relational neurons}\label{fig:example}
\end{figure}

\smallskip
\begin{example}\label{ex:relSplitting}
In Fig.~\ref{fig:example}, we depict a process of verifying a tiny neural network by an approximation verifier (Fig.~\ref{fig:apprx_propagation}), and then refining the approximated bound of the output difference by problem splitting (Fig.~\ref{fig:after_split}).  

In particular, by splitting an individual neuron $\indNeuronPre{1}{1}$, the bound of the difference $\relNeuronPre{2}{1}$ between two outputs of two individual inferences is refined to be $[-0.15, 0.15]$.

In contrast, by branching on the relational neuron $\relNeuronPre{1}{1}$, the bound of $\relNeuronPre{2}{1}$ is refined to be $[-0.067, 0.15]$, which is tighter than the result of splitting individual neurons. \hfill$\lhd$
\end{example}

\smallskip
As observed in Example~\ref{ex:relSplitting}, splitting individual neurons in relational verification may lead to suboptimal performance, compared to splitting relational neuron, and therefore may not be the best choice in the context of relational verification.
Intuitively, in relational verification, as the differences between individual neurons matter, splitting individual neurons may not be as helpful as directly splitting relational neurons in narrowing down the difference bounds between individual neurons. In light of this, we propose a \bab approach for relational verification, which features problem splitting via relational neurons.

\section{The Proposed \bab Approach}\label{sec:approach}
By~\S{}\ref{sec:motivations}, it is desirable to refine approximation in relational verification, by using \bab with relational neuron splitting. In this section, we detail this approach, with a focus on the selection of relational neurons in~\S{}\ref{sec:rel_neuron_selection}.

\subsection{Approach Overview}\label{sec:appOverview}

\myparagraph{Sub-problem construction} We first define \emph{sub-problems} in our context. A verification problem defined in Def.~\ref{def:problemStatement} is identified by a given neural network and a global robustness property. We generalize this problem by introducing \emph{relational constraints} $\relspec$ predicating over $\relNeuronPre{i}{j} = \indNeuronPre{i}{j} - \indNeuronTwoPre{i}{j}$, i.e., the inputs of the non-linear transformation in relational neurons.

\begin{definition}[Relational constraints]
    Given a pair $(i,j)$, a relational constraint $\varphi$ w.r.t. $(i,j)$ is a proposition either $\relNeuronPre{i}{j}\le 0$ (written as $\negRel{\relNeuronPre{i}{j}}$) or $\relNeuronPre{i}{j}\ge 0$ (written as $\posRel{\relNeuronPre{i}{j}}$). Multiple relational constraints, each w.r.t. a different $(i,j)$, can form a conjunction $\relspec = \varphi_1\land \cdots \land \varphi_{\relConLen}$, which we call a relational constraint sequence. $\relspec$ can be empty, in which case we denote it by $\top$. 
\end{definition}

\begin{definition}[(Sub-)problems of relational verification]\label{def:subProblem}
    Given a neural network \network, a global robustness property identified by $(\inpSpace, \inpSpecDist, \outSpecDist)$ (see Def.~\ref{def:glb_rbst}), and a relational constraint sequence $\relspec$, a verification problem is to determine whether $\network$ satisfies global robustness under the constraint of $\relspec$.    
\end{definition}

\medskip
We also generalize the approximation verifier $\appVerifierRel$ introduced in~\S{}\ref{sec:OverAppVerification}, by allowing it to handle an additional argument $\relspec$. 
Notably, an originally given verification problem in Def.~\ref{def:problemStatement}  can be seen as a special case of Def.~\ref{def:subProblem}, by treating $\relspec$ of the original problem as $\top$; consequently, we can apply $\appVerifierRel$ not only to the original problem but also to its sub-problems. While there can be different ways of accommodating this additional constraint by \appVerifierRel, below we introduce a linear programming (LP)-based approach, which allows to obtain tight relational bounds.

In our problem, since we aim to show that the difference $|\relNeuronPost{\outLayerIdx}{\labelIdx}| \le \outSpecDist$, we need to consider both the maximum and the minimum of $\relNeuronPost{\outLayerIdx}{\labelIdx}$. Here we take the minimization of $\relNeuronPost{\outLayerIdx}{\labelIdx}$ as an example to showcase our approach; the maximization of $\relNeuronPost{\outLayerIdx}{\labelIdx}$ can be computed similarly. The LP program we need to solve is as follows, subject to six categories of constraints:
\begin{equation}
    \begin{aligned}
 \min \quad &\relNeuronPost{\outLayerIdx}{\labelIdx} \label{eq:rel_prim_obj} \\
 \text{s.t.} \quad &\constInput, \quad \constAffine, \quad \constRelAffine, \\
 & \constIndRelu, \quad \constRelRelu, \quad \relspec
\end{aligned}
\end{equation}
where \constInput denotes the constraints related to the input layer, including that for individual inferences and for the difference between the two inferences:
\begin{compactitem}
    \item $\indNeuronOne{0}{} \geq \inpSpaceLower{}$, $\indNeuronOne{0}{} \leq \inpSpaceUpper{}$, $\indNeuronTwo{0}{} \geq \inpSpaceLower{}$, $\indNeuronTwo{0}{} \leq \inpSpaceUpper{}$
    \item $\relNeuronPost{0}{} \geq -\inpSpecDist$, $\relNeuronPost{0}{} \leq \inpSpecDist$,
\end{compactitem}
\constAffine and \constRelAffine respectively denote the constraints associated with the affine transformation of two individual and relational neurons:
\begin{compactitem}
    \item $x\uid{i+1} = W\uid{i}y\uid{i} + b\uid{i}, \; \forall i \in \{0, \ldots, \outLayerIdx-1\}$
    \item $\hat{x}\uid{i+1} = W\uid{i}\hat{y}\uid{i} + b\uid{i}, \; \forall i \in \{0, \ldots, \outLayerIdx-1\}$
    \item $\relNeuronPre{i+1}{} = W\uid{i}\relNeuronPost{i}{}, \; \forall i \in \{0, \ldots, \outLayerIdx-1\}$,
\end{compactitem}
\constIndRelu denotes the constraints for \relu functions in each individual neuron:
\begin{compactitem}
    \item $\indNeuron{i}{j} = 0, \; \forall i \in \{1, \ldots, \outLayerIdx-1\},  \; j \in \reluNeg_i$
    \item $\indNeuron{i}{j} = \indNeuronPre{i}{j}, \; \forall i \in \{1, \ldots, \outLayerIdx-1\}, \; j \in \reluPos_i$
    \item $\indNeuronTwoPost{i}{j} = 0, \; \forall i \in \{1, \ldots, \outLayerIdx-1\}, \; j \in \reluNegHat_i$
    \item $\indNeuronTwoPost{i}{j} = \indNeuronTwoPre{i}{j}, \; \forall i \in \{1, \ldots, \outLayerIdx-1\}, \; j \in \reluPosHat_i$
\item $\left.\begin{aligned}
  &\indNeuron{i}{j} \geq \slope{i}{j} \indNeuronPre{i}{j}\\
  &(\upperBound{\indNeuronPre{i}{j}} - \lowerBound{\indNeuronPre{i}{j}})\indNeuron{i}{j} - \\ & \qquad ( \upperBound{\indNeuronPre{i}{j}} \indNeuronPre{i}{j} - \upperBound{\indNeuronPre{i}{j}} \lowerBound{\indNeuronPre{i}{j}} ) \leq 0
\end{aligned}\right\} 
  \begin{aligned}
    &\forall i \in \{1, \ldots, \outLayerIdx-1\} \\
    &j \in \reluUns_i
\end{aligned}
$
    \item $\left.\begin{aligned}
  &\indNeuronTwoPost{i}{j} \geq \slopeHat{i}{j} \indNeuronTwoPre{i}{j}\\
  &(\upperBound{\indNeuronTwoPre{i}{j}} - \lowerBound{\indNeuronTwoPre{i}{j}})\indNeuronTwoPost{i}{j} -  \\ & \qquad ( \upperBound{\indNeuronTwoPre{i}{j}} \hat{x}\uid{i} - \upperBound{\indNeuronTwoPre{i}{j}} \lowerBound{\indNeuronTwoPre{i}{j}} ) \leq 0
\end{aligned}\right\} \begin{aligned}
    & \forall i \in \{1, \ldots, \outLayerIdx-1\}, \\
    & j \in \reluUnsHat_i,
\end{aligned}$
\end{compactitem}
where the bounds $[\lowerBound{\indNeuronPre{i}{j}}, \upperBound{\indNeuronPre{i}{j}}]$ are obtained by the previous layers, and $\reluNeg_i$, $\reluPos_i$ and $\reluUns_i$ respectively represent the set of indexes $j$ for which the ReLU input $\indNeuronPre{i}{j}$ is negative, positive, or undecided, as follows; similarly, $\reluNegHat_i$, $\reluPosHat_i$ and $\reluUnsHat_i$ represent the cases for the corresponding individual neuron.
\begin{compactitem}
    \item $\reluNeg_i = \{ j\mid \upperBound{\indNeuronPre{i}{j}} \le 0 \}$, 
    \item $\reluPos_i = \{ j\mid \lowerBound{\indNeuronPre{i}{j}} \ge 0 \}$, 
    \item $\reluUns_i = \{ j\mid \lowerBound{\indNeuronPre{i}{j}} < 0 < \upperBound{\indNeuronPre{i}{j}} \}$
\end{compactitem}
\constRelRelu denotes the constraints of \relu function for relational neurons based on the abstraction domain shown in Fig.~\ref{fig:diffpoly}. The propagation is similar to that for individual neurons, and has been detailed in~\S{}\ref{sec:OverAppVerification}. In practice, we also involve the heuristics imposed by \raven, which leverages the bounds of individual neurons to tighten the bounds of relational bounds. Due to the page limit, we do not expand the whole approach here, but we leave it in Appendix~\ref{appendix:relu_transformation_raven}.

The last constraints \relspec, i.e., relational constraints, are derived from problem splitting. As these are also linear constraints, LP solvers can handle them effectively.

\myparagraph{Algorithm} Similar to classic \bab, the proposed \emph{relational \bab} also performs verification by iterative problem splitting and application of approximation verifiers, such as \raven~\cite{banerjee24raven}, to the sub-problems.

\begin{algorithm*}[!tb]
\caption{The proposed \bab for relational verification}
\label{alg:relational_bab}
\small
\begin{algorithmic}[1]
\Require A neural network $\network$, global robustness identified by $(\inpSpecDist, \outSpecDist)$ and $\inpSpace$

\State $\queue \gets \{\top\}$ \label{line:initQ} \Comment{initialize \queue to record problems to be solved}
\State $\Call{RelBaB}{\network, \inpSpecDist, \outSpecDist, \inpSpace, \queue}$ \label{line:startBaB} \Comment{invoke the verification process}
\Function{RelBaB}{$\network, \inpSpecDist, \outSpecDist, \inpSpace, \queue$}
    \If{\Call{Empty}{\queue}} \label{line:checkEmpty}
        \State \Return \TRUE \label{line:returnTrue}
    \EndIf
    \State $\relspec\gets \Call{Pop}{\queue}$ \label{line:popQ} \Comment{select a sub-problem}
    \State $\lowerBound{\relNeuronPost{i}{j}}, \upperBound{\relNeuronPost{i}{j}}, \lowerBound{\indNeuronOne{i}{j}}, \upperBound{\indNeuronOne{i}{j}}, \lowerBound{\indNeuronTwo{i}{j}}, \upperBound{\indNeuronTwo{i}{j}}, \left(\cex{\indNeuronOne{0}{}}, \cex{\indNeuronTwo{0}{}}\right) \gets \appVerifierRel(\network, \inpSpace, \inpSpecDist, \relspec)$
\label{line:callVerifier}    \If{$\upperBound{\relNeuronPost{\outLayerIdx}{j}} > \outSpecDist$ or $\lowerBound{\relNeuronPost{\outLayerIdx}{j}} < -\outSpecDist$} \label{line:failVerify} \Comment{not verified}
        \If{$\left|\network(\cex{\indNeuronOne{0}{}}) - \network(\cex{\indNeuronTwo{0}{}})\right| > \outSpecDist$} \label{line:validated} \Comment{real counterexample}
            \State \Return\FALSE \label{line:returnFalse}
        \Else
            \State $(i, j) \gets \Call{RelSelect}{
            \network, \inpSpecDist, \outSpecDist, \inpSpace, \lowerBound{\relNeuronPost{i}{j}}, \upperBound{\relNeuronPost{i}{j}}, \lowerBound{\indNeuronOne{i}{j}}, \upperBound{\indNeuronOne{i}{j}}, \lowerBound{\indNeuronTwo{i}{j}}, \upperBound{\indNeuronTwo{i}{j}}
            }$ \label{line:selectNeuron}
            \For{$\varphi\in \left\{\posRel{\relNeuronPre{i}{j}}, \negRel{\relNeuronPre{i}{j}}\right\}$} \label{line:subProblem}
                \State $\Call{Push}{\queue, \relspec\land \varphi}$ \label{line:pushQ} \Comment{push to \queue for future verification}
            \EndFor
        \EndIf
    \EndIf
    \State \Return $\Call{RelBaB}{\network, \inpSpecDist, \outSpecDist, \inpSpace, \queue}$ \label{line:recursiveCall} \Comment{recursive call}
\EndFunction

\end{algorithmic}
\end{algorithm*}

Alg.~\ref{alg:relational_bab} presents the workflow of our \bab approach. It adopts a queue $\queue$ to record the (sub-)problems yet to be checked, which is initialized to contain the original verification problem, identified by $\top$ only (Line~\ref{line:initQ}). Then, it goes through the loop of problem splitting and bounding (Line~\ref{line:startBaB}).   

In function \textsc{RelBaB}, a given problem in $\queue$ is checked by approximation verifiers and, if necessary, split to sub-problems for further processing. If no problem is in \queue, it signifies that all (sub-)problems have been solved, so verification can be terminated and return \TRUE (Line~\ref{line:returnTrue}). Otherwise, a sub-problem, identified by a relational constraint sequence $\relspec$, will be popped from \queue and  checked by the approximation verifier \appVerifierRel (Line~\ref{line:popQ}), using the aforementioned approach. By applying \appVerifierRel to the sub-problem $\relspec$, \appVerifierRel can return the bounds for both relational neurons and individual neurons (Line~\ref{line:callVerifier}), by which we can check whether the global robustness is satisfied. If not satisfied, we can validate the counterexample $\left(\cex{\indNeuronOne{0}{}}, \cex{\indNeuronTwo{0}{}}\right)$ returned by \appVerifierRel by feeding it into \network (Line~\ref{line:validated}), and terminate the verification by returning \FALSE if the counterexample is a real one (Line~\ref{line:returnFalse}). 

If the counterexample is spurious, we split the problem to sub-problems by adding new relational constraints $\relspec$. Unlike conventional \bab approaches~\cite{bunel2020branch,suresh24rabbit} that branch on individual neurons, our approach performs branching on \emph{relational neurons}. While it is possible to consider both individual and relational neurons as candidates for splitting, currently in our approach we select relational neurons exclusively, because given a  verification problem reasonably difficult, there are often sufficiently many relational neurons to split within a limited time budget, and so they are preferred to be selected to maximize the refinement.
 Specifically, we first select a relational neuron identified by its index $(i,j)$ (Line~\ref{line:selectNeuron}), and then add the corresponding propositions, i.e., $\posRel{\relNeuronPre{i}{j}}$  and $\negRel{\relNeuronPre{i}{j}}$ to $\relspec$ to construct new sub-problems (Line~\ref{line:subProblem}). These two new sub-problems will be added to \queue such that they can be later accessed (Line~\ref{line:pushQ}). Then, Alg.~\ref{alg:relational_bab} recursively calls \textsc{RelBaB} with the updated \queue to proceed with the verification of sub-problems (Line~\ref{line:recursiveCall}). 

It remains an important question in Alg.~\ref{alg:relational_bab} regarding the selection of relational neurons, i.e., the implementation of \textsc{RelSelect} in Line~\ref{line:selectNeuron}. Similarly to classic \bab, it is of great significance to select a proper neuron each time that can maximize the refinement of abstraction, thereby solving the problem efficiently. In~\S{}\ref{sec:rel_neuron_selection}, we elaborate on our proposed neuron selection strategy.

\subsection{Relational Neuron Selection}\label{sec:rel_neuron_selection}
As reviewed in~\S{}\ref{sec:bab}, mainstream \bab-based approaches \cite{wang2021beta,bunel2020branch,suresh24rabbit} leverage the dual formulation of verification problems for efficient neuron selection. 
However, this approach cannot be directly used to select relational neurons in \bab for relational verification. 
To bridge this gap, we extend the neuron selection approach \babsr~\cite{bunel2020branch} in classic verification to the relational settings, based on the dual formulation of the LP program in~\S{}\ref{sec:appOverview}.
Similarly to \babsr, the objective function of our dual formulation is also a summation of neuron-wise terms, so we can use them to estimate the bounds of relational neurons in the output layer efficiently.

\myparagraph{Relational Dual Formulation}
In general, to obtain the dual problem of a given primal problem, we need to go through two steps: 
\begin{inparaenum}[1)]
    \item construct a Lagrangian function by adding Lagrangian multipliers to constraints in the primal problem; and
    \item construct the dual problem by using the Karush-Kuhn-Tucker (KKT) condition.
\end{inparaenum}

In our context, the primal problem is the linear program formulated in Eq.~\ref{eq:rel_prim_obj}. For clarity, we take the formulation for the original problem where $\relspec = \top$ to illustrate our approach, while the dual problem can be updated  as Alg.~\ref{alg:relational_bab} progresses. 
Below, we go through the two steps mentioned above to obtain the dual problem. 
Due to the space limit, the full derivation including technical details are provided in Appendix~\ref{appendix:rel_dual}. 
As a result, the derived dual problem is expressed as follows:
\begin{align}
& \max \notag \\
& \begin{aligned}
& \inpSpaceLower{} \left[ \preReluInd[0] \right]_- - \inpSpaceUpper{} \left[ \preReluInd[0] \right]_+ 
+ \inpSpaceLower{} \left[ \preReluIndHat[0] \right]_- - \inpSpaceUpper{} \left[ \preReluIndHat[0] \right]_+\\ 
&- \inpSpecDist \left[ \preReluDelta[0] \right]_- - \inpSpecDist \left[ \preReluDelta[0] \right]_+\\
\end{aligned} \label{eq:dual_obj_inp} \\
&- \sum_{i=0}^{\outLayerIdx-1} \bigl( (\affineInd[i+1] + \affineIndHat[i+1])\transposed b\uid{i} \bigr) \label{eq:dual_obj_affine} \\
&+ \sum_{i=1}^{\outLayerIdx-1} \sum_{j} \begin{aligned}\bigl(& \objCoefInd{i}{j} \left[ \preReluInd[i]_j \right]_+ + \objCoefIndHat{i}{j} \left[ \preReluIndHat[i]_j \right]_+ \\
& + \objCoefDeltaPos{i}{j} \left[ \preReluDelta[i]_j \right]_+ + \objCoefDeltaNeg{i}{j} \left[ \preReluDelta[i]_j \right]_- \bigr)\end{aligned} \label{eq:dual_obj_relu}
\end{align}
where the operators $[z]_- = \max(0, -z)$ and $[z]_+ = \max(0, z)$ are defined for a given $z\in\R$, and $\affineInd$, $\affineIndHat$, $\affineDelta$, $\preReluInd$, $\preReluIndHat$ and $\preReluDelta$ are the Lagrangian multipliers (in dual context, we call them dual variables). These dual variables $\affineInd$, $\affineIndHat$, $\affineDelta$, $\preReluInd$, $\preReluIndHat$ and $\preReluDelta$ are derived as constants by backpropagation from the last layer to input layer:
{\setlength{\abovedisplayskip}{6pt}
 \setlength{\belowdisplayskip}{6pt}
\begin{align}
&\affineInd[\outLayerIdx] = \affineIndHat[\outLayerIdx] = 0, ~ \affineDelta[\outLayerIdx] = -C \label{eq:rel_dual_const_output} \\
&\left.
\begin{aligned}
&\preReluInd[i] = W^{(i)}\transposed \affineInd[i+1],\\ 
& \preReluIndHat[i] = W^{(i)}\transposed \affineIndHat[i+1],~ \\
&\preReluDelta[i] = W^{(i)}\transposed \affineDelta[i+1], 
\end{aligned} \right\}
\label{eq:rel_dual_const_affine} \quad \forall i \in \{\outLayerIdx-1, \ldots, 0\} \\
&\left.
\begin{aligned}
&\affineInd[i]_j = \constCoefInd{i}{j} \preReluInd[i]_j + \constCoefIndDelta{i}{j} \preReluDelta[i]_j \\
&\affineIndHat[i]_j = \constCoefIndHat{i}{j} \preReluIndHat[i]_j + \constCoefIndHatDelta{i}{j} \preReluDelta[i]_j \\
&\affineDelta[i]_j = \constCoefDeltaPos{i}{j} \left[ \preReluDelta[i]_j \right]_+ + \constCoefDeltaNeg{i}{j} \left[ \preReluDelta[i]_j \right]_-\\
\end{aligned}\right\}\; \begin{aligned}
& \forall i \in \{ \outLayerIdx-1, \\
& \; \ldots, 1 \} 
\end{aligned} \label{eq:rel_dual_const_relu}
\end{align}}
where $C$ is a vector whose $\labelIdx$-th element is $1$ and all other elements are $0$ to point out the target dimensional output relational bound (i.e., $C$ is a one-hot vector based on the global robustness specification in Def.~\ref{def:glb_rbst}), 
and the coefficients $\constCoefInd{i}{j}$, $\constCoefIndDelta{i}{j}$, $\constCoefIndHat{i}{j}$, $\constCoefIndHatDelta{i}{j}$, $\constCoefDeltaPos{i}{j}$, $\constCoefDeltaNeg{i}{j}$, $\objCoefInd{i}{j}$, $\objCoefIndHat{i}{j}$, $\objCoefDeltaPos{i}{j}$, and $\objCoefDeltaNeg{i}{j}$ in Eq.~(\ref{eq:dual_obj_inp}, \ref{eq:dual_obj_affine}, \ref{eq:dual_obj_relu}) and Eq.~(\ref{eq:rel_dual_const_output}, \ref{eq:rel_dual_const_affine}, \ref{eq:rel_dual_const_relu}) are derived based on the conditions of \relu inputs of individual and relational neurons, as shown in Table~\ref{tab:coef}.

\begin{table*}[!tb]
\centering
\begin{threeparttable}
\centering
\caption{Coefficients in dual formulation. In the column ``case'', \reluStateOne, \reluStateTwo, and \reluStateRel, together with their sign notations, denote the sets of individual and relational neurons categorized according to their pre-activation conditions (details of the categorization are provided in the footnote of the table). This table presents the correspondence between the symbols appearing in the dual formulation (specifically, Eq.~\ref{eq:dual_obj_relu} and Eq.~\ref{eq:rel_dual_const_relu}) and the coefficients in Eq.~\ref{eq:coef} for each combination of the pre-activation conditions.}
\label{tab:coef}
\setlength{\tabcolsep}{10pt}
\begin{tabular}{c|c|c|c|c|c|c|c|c|c|c}
\toprule
case &
$\constCoefInd{i}{j}$ &
$\constCoefIndDelta{i}{j}$ &
$\constCoefIndHat{i}{j}$ &
$\constCoefIndHatDelta{i}{j}$ &
$\constCoefDeltaPos{i}{j}$ &
$\constCoefDeltaNeg{i}{j}$ &
$\objCoefInd{i}{j}$ &
$\objCoefIndHat{i}{j}$ &
$\objCoefDeltaPos{i}{j}$ &
$\objCoefDeltaNeg{i}{j}$ \\
\midrule
$j \in \reluNeg_i \cap \reluNegHat_i$ & 0 & 0 & 0 & 0 & 0 & 0 & 0 & 0 & 0 & 0 \\
$j \in \reluPos_i \cap \reluNegHat_i$ & 1 & 1 & 0 & 0 & 0 & 0 & 0 & 0 & 0 & 0 \\
$j \in \reluNeg_i \cap \reluPosHat_i$ & 0 & 0 & 1 & -1 & 0 & 0 & 0 & 0 & 0 & 0 \\
$j \in \reluPos_i \cap \reluPosHat_i$ & 1 & 1 & 1 & -1 & 0 & 0 & 0 & 0 & 0 & 0 \\
$j \in \reluUns_i \cap \reluNegHat_i$ & $\constCmplxInd{i}{j}$ & $\constCmplxInd{i}{j}$ & 0 & 0 & 0 & 0 & $\objCmplxInd{i}{j}$ & 0 & $\objCmplxInd{i}{j}$ & 0 \\
$j \in \reluNeg_i \cap \reluUnsHat_i$ & 0 & 0 & $\constCmplxIndHat{i}{j}$ & $\constCmplxIndHat{i}{j}$ & 0 & 0 & 0 & $\objCmplxIndHat{i}{j}$ & 0 & $\objCmplxIndHat{i}{j}$ \\
$j \in \reluUns_i \cap \reluPosHat_i$ & $\constCmplxInd{i}{j}$ & $\constCmplxInd{i}{j}$ & 1 & -1 & 0 & 0 & $\objCmplxInd{i}{j}$ & 0 & $\objCmplxInd{i}{j}$ & 0 \\
$j \in \reluPos_i \cap \reluUnsHat_i$ & 1 & 1 & $\constCmplxIndHat{i}{j}$ & $\constCmplxIndHat{i}{j}$ & 0 & 0 & 0 & $\objCmplxIndHat{i}{j}$ & 0 & $\objCmplxIndHat{i}{j}$ \\
$j \in \reluUns_i \cap \reluUnsHat_i \cap \relReluPos_i$ & $\constCmplxInd{i}{j}$ & $\constCmplxInd{i}{j}$ & $\constCmplxIndHat{i}{j}$ & $\constCmplxIndHat{i}{j}$ & 1 & 0 & $\objCmplxInd{i}{j}$ & $\objCmplxIndHat{i}{j}$ & 0 & 0 \\
$j \in \reluUns_i \cap \reluUnsHat_i \cap \relReluNeg_i$ & $\constCmplxInd{i}{j}$ & $\constCmplxInd{i}{j}$ & $\constCmplxIndHat{i}{j}$ & $\constCmplxIndHat{i}{j}$ & 0 & -1 & $\objCmplxInd{i}{j}$ & $\objCmplxIndHat{i}{j}$ & 0 & 0 \\
$j \in \reluUns_i \cap \reluUnsHat_i \cap \relReluUns_i$ & $\constCmplxInd{i}{j}$ & $\constCmplxInd{i}{j}$ & $\constCmplxIndHat{i}{j}$ & $\constCmplxIndHat{i}{j}$ & $\constCmplxDeltaU{i}{j}$ & $\constCmplxDeltaL{i}{j}$ & $\objCmplxInd{i}{j}$ & $\objCmplxIndHat{i}{j}$ & $\objCmplxIndDelta{i}{j}$ & $\objCmplxIndDelta{i}{j}$ \\ \bottomrule
\end{tabular}
\begin{tablenotes}
\scriptsize
    \item[$\ast$] $\reluNeg_i = \{ j\mid \upperBound{\indNeuronPre{i}{j}} \le 0 \}$, $\reluPos_i = \{ j\mid \lowerBound{\indNeuronPre{i}{j}} \ge 0 \}$, $\reluUns_i = \{ j\mid \lowerBound{\indNeuronPre{i}{j}} < 0 < \upperBound{\indNeuronPre{i}{j}} \}$
    \item[$\dagger$] $\reluNegHat_i = \{ j\mid \upperBound{\indNeuronTwoPre{i}{j}} \le 0 \}$, $\reluPosHat_i = \{ j\mid \lowerBound{\indNeuronTwoPre{i}{j}} \ge 0 \}$, $\reluUnsHat_i = \{ j\mid \lowerBound{\indNeuronTwoPre{i}{j}} < 0 < \upperBound{\indNeuronTwoPre{i}{j}} \}$
    \item[$\ddagger$] $\relReluNeg_i = \{ j\mid \upperBound{\relNeuronPre{i}{j}} \le 0 \}$, $\relReluPos_i = \{ j\mid \lowerBound{\relNeuronPre{i}{j}} \ge 0 \}$, $\relReluUns_i = \{ j\mid \lowerBound{\relNeuronPre{i}{j}} < 0 < \upperBound{\relNeuronPre{i}{j}} \}$
\end{tablenotes}
\end{threeparttable}
\end{table*}
In Table~\ref{tab:coef}, to group the \relu input conditions of the individual and relational neurons, we use three categories $\reluStateOne$, $\reluStateTwo$, and $\reluStateRel$, as annotated in the table. These represent the sets of indexes $j$ grouped according to the pre-activation states of $\indNeuronPre{i}{j}$, $\indNeuronTwoPre{i}{j}$, and $\relNeuronPre{i}{j}$ (e.g., $j \in \reluUnsHat_i$ if $\lowerBound{\indNeuronTwoPre{i}{j}} < 0 < \upperBound{\indNeuronTwoPre{i}{j}}$). The symbols in Table~\ref{tab:coef} are computed by the pre-activation bounds:
\begin{align}
& \constCmplxInd{i}{j} = \frac{\overline{\indNeuronPre{i}{j}}}{\overline{\indNeuronPre{i}{j}} - \underline{\indNeuronPre{i}{j}}}, \;
\constCmplxIndHat{i}{j} = \frac{\overline{\indNeuronTwoPre{i}{j}}}{\overline{\indNeuronTwoPre{i}{j}} - \underline{\indNeuronTwoPre{i}{j}}}, \;
\objCmplxInd{i}{j} = \frac{\overline{\indNeuronPre{i}{j}} \underline{\indNeuronPre{i}{j}}}{\overline{\indNeuronPre{i}{j}} - \underline{\indNeuronPre{i}{j}}}, \notag \\
& \objCmplxIndHat{i}{j} = \frac{\overline{\indNeuronTwoPre{i}{j}} \underline{\indNeuronTwoPre{i}{j}}}{\overline{\indNeuronTwoPre{i}{j}} - \underline{\indNeuronTwoPre{i}{j}}}, \quad
\constCmplxDeltaU{i}{j} = \frac{\overline{\relNeuronPre{i}{j}}}{\overline{\relNeuronPre{i}{j}} - \underline{\relNeuronPre{i}{j}}}, \label{eq:coef} \\
& \constCmplxDeltaL{i}{j} = \frac{\underline{\relNeuronPre{i}{j}}}{\overline{\relNeuronPre{i}{j}} - \underline{\relNeuronPre{i}{j}}}, \quad
\objCmplxIndDelta{i}{j} =  \frac{\overline{\relNeuronPre{i}{j}} \underline{\relNeuronPre{i}{j}}}{\overline{\relNeuronPre{i}{j}} - \underline{\relNeuronPre{i}{j}}} \notag
\end{align}


In the objective of dual problem, Eq.~\ref{eq:dual_obj_inp}, \ref{eq:dual_obj_affine}, and \ref{eq:dual_obj_relu} respectively correspond to the input constraints \constInput, affine transformation constraints of individual neurons \constAffine, and \relu transformation constraints of individual and relational neurons \constIndRelu and \constRelRelu. For dual constraints Eq.~\ref{eq:rel_dual_const_output}, \ref{eq:rel_dual_const_affine}, and \ref{eq:rel_dual_const_relu}, they are constructed based on the \constAffine, \constRelAffine, \constIndRelu, and \constRelRelu through the backward path of the network, while also collecting dual variables from output layer to input layer (e.g., $\affineInd[\outLayerIdx], \affineIndHat[\outLayerIdx], \affineDelta[\outLayerIdx], \ldots ,\preReluInd[0], \preReluIndHat[0], \preReluDelta[0]$).

Consequently, this objective function also consists of the terms, each of which represents the neuron-wise contribution, and so can be used to estimate the impact of neuron splitting, as classic \babsr does.

\myparagraph{Neuron selection}
Using the dual formulation in Eq.~\ref{eq:dual_obj_inp}, \ref{eq:dual_obj_affine}, \ref{eq:dual_obj_relu}, \ref{eq:rel_dual_const_output}, \ref{eq:rel_dual_const_affine}, \ref{eq:rel_dual_const_relu}, we estimate the improvement in the bounds of relational neurons in the output layer, resulting from splitting a relational neuron. Note that, for the dual objective in Eq.~\ref{eq:dual_obj_inp}-\ref{eq:dual_obj_relu}, the splitting affects the relevant parts in Eq.~\ref{eq:dual_obj_affine} and Eq.~\ref{eq:dual_obj_relu}; in the following, we elaborate on how the value of each part in this dual objective function is affected by neuron splitting.


In line with classic \babsr, the high-level idea is to estimate the change of the value of the objective function (Eq.~\ref{eq:dual_obj_inp}-\ref{eq:dual_obj_relu}) before and after splitting. Given a specific relational neuron $\relNeuronPre{i}{j}$ with index $(i,j)$, splitting $\relNeuronPre{i}{j}$ can lead to the corresponding changes in $\indNeuronPre{i}{j}$ and $\indNeuronTwoPre{i}{j}$ (due to the relation $\relNeuronPre{i}{j} = \indNeuronPre{i}{j} - \indNeuronTwoPre{i}{j}$), which thus further affects the activation conditions of the corresponding ReLUs at $(i,j)$. By Table~\ref{tab:coef}, the activation conditions of a ReLU affects the values of 
the coefficients $\constCoefInd{i}{j}$, $\constCoefIndDelta{i}{j}$, $\constCoefIndHat{i}{j}$, $\constCoefIndHatDelta{i}{j}$, $\constCoefDeltaPos{i}{j}$, $\constCoefDeltaNeg{i}{j}$, $\objCoefInd{i}{j}$, $\objCoefIndHat{i}{j}$, $\objCoefDeltaPos{i}{j}$, and $\objCoefDeltaNeg{i}{j}$, so it yields to a change of the values of these coefficients. Then, these coefficients affect Eq.~\ref{eq:rel_dual_const_relu}, which finally affect the change of Eq.~\ref{eq:dual_obj_affine} and Eq.~\ref{eq:dual_obj_relu} in the objective function of the dual problem.

Below, we denote the variables with prime (e.g., $\affineIndPrime[i]_j, \objCoefIndPrime{i}{j}$) as the updated dual variables or coefficients after splitting. By applying this change to Eq.~\ref{eq:rel_dual_const_relu}, we obtain the updated $\affineIndPrime[i]_j$ and $\affineIndHatPrime[i]_j$, then the refinement of Eq.~\ref{eq:dual_obj_affine} in the dual objective is expressed as follows:
\begin{equation}\label{eq:final_1}
\left( \affineInd[i]_j + \affineIndHat[i]_j \right) b\uid{i-1}_j - \left( \affineIndPrime[i]_j + \affineIndHatPrime[i]_j \right) b\uid{i-1}_j
\end{equation}
Moreover, as the changes of the coefficients also affect Eq.~\ref{eq:dual_obj_relu} in the dual objective, so it also brings the following refinement:
{\mathcompact
\begin{equation}\label{eq:final_2}
\begin{aligned}
& - \Bigl( \objCoefInd{i}{j} \left[ \preReluInd[i]_j \right]_+ + \objCoefIndHat{i}{j} \left[ \preReluIndHat[i]_j \right]_+ + \objCoefDeltaPos{i}{j} \left[ \preReluDelta[i]_j \right]_+ \\
& + \objCoefDeltaNeg{i}{j} \left[ \preReluDelta[i]_j \right]_- \Bigr) + \Bigl(  \objCoefIndPrime{i}{j} \left[ \preReluInd[i]_j \right]_+ +  \objCoefIndHatPrime{i}{j} \left[ \preReluIndHat[i]_j \right]_+ \\
& + \objCoefDeltaPosPrime{i}{j} \left[ \preReluDelta[i]_j \right]_+ + \objCoefDeltaNegPrime{i}{j} \left[ \preReluDelta[i]_j \right]_- \Bigr)
\end{aligned}
\end{equation}}

As a result, the total refinement can be considered as the aggregation between these two parts, namely, it can be obtained by taking the sum of Eq.~\ref{eq:final_1} and Eq.~\ref{eq:final_2}. 

In line with the classic \babsr, the computation of such refinement estimation can be done very efficiently in a constant time complexity, because at each step, we already know the values of all the terms necessary to compute the values of Eq.~\ref{eq:final_1} and Eq.~\ref{eq:final_2}.


\section{Experimental Evaluation}\label{sec:experiment}
We evaluate the performance of \tool for  verification problems against global robustness (Def.~\ref{def:glb_rbst}) across various datasets and network architectures. Our code and data are publicly available~\cite{SupplementaryMaterialEMSOFT2026}.

\subsection{Experiment Settings}\label{sec:baselines}
\myparagraph{Evaluation Baselines}
To evaluate \tool, we compare it against \raven~\cite{banerjee24raven}, 
which is a recent approximation-based approach targeting relational verification. Moreover, our evaluation also covers \rabbit~\cite{suresh24rabbit}, which is the only relational verification approach that utilizes \bab, to the best of our knowledge. However, \rabbit is not directly applicable to our setting because it targets a different relational property from ours. To compare with it, we extract the essential strategy of \rabbit for problem splitting, and adapt it to our setting. Additionally, we also include ablation variants of \tool, \is and \RSrandom, in our evaluation. Details of the baselines are provided as follows:
\begin{compactitem}[$\bullet$]
\item \raven~\cite{banerjee24raven}: A state-of-the-art approximation verifier without abstraction refinement, as introduced in~\S{}\ref{sec:OverAppVerification}. By comparison with \raven, we can gain an insight into the effectiveness of the refinement brought by \bab. 

\item \baseline: A \bab approach based on problem splitting of \emph{individual} neurons. It also adopts the \babsr-style selection, but unlike \tool, it does not consider relational neurons, but instead, it solely relies on individual inferences (as classic \babsr does), by which it assesses the impact of individual neuron splitting to the final output bounds.
This is somehow in line with the strategy  of \rabbit~\cite{suresh24rabbit}, the only work using \bab for relational verification, as far as we know. We cannot directly compare with \rabbit, because it targets a different specification called \emph{universal adversarial perturbations}, which derives a different problem setting from ours; nevertheless, our comparison with \baseline sheds light on the differences, because \baseline adopts essentially a similar problem splitting strategy with \rabbit.
 \item \is: Similar to \baseline,  \is splits individual neurons. Unlike \baseline, \is relies on our dual formulation that considers bound propagation for both individual neurons and relational neurons, and selects individual neurons, rather than relational neurons as \tool does, to split. 
 \item \RSrandom:  An ablation variant of \tool that retains the relational splitting framework but substitutes the proposed dual-based selection strategy with a uniform random selection from the candidate neurons.
\end{compactitem}

\myparagraph{Metrics} We evaluate performance using the following metrics on \emph{efficiency}, and \emph{bound refinement} of each approach:

\begin{compactitem}[$\bullet$]
    \item Number of solved instances (\solved): the number of verification problems successfully resolved within the time budget, serving as the primary indicator of verification capability. For all \bab-based approaches (i.e., \baseline, \is, \RSrandom and \tool), \solved counts only the instances additionally solved beyond \raven, since all \bab methods invoke \raven first and branch only on instances \raven fails to verify. 
    \item Number of sub-problems (\subproblems): the average number of sub-problems explored per solved instance, measuring search efficiency, i.e., fewer splits for the same verdict indicates a more effective branching strategy.
    \item Time ratio (\deltaTime): as we assign different time budget for different datasets (due to their different complexities), we represent verification time as a percentage of the time budget, i.e., \deltaTime $= (t / t_{\text{budget}}) \times 100\%$, where $100\%$ denotes a timeout. In RQ2 and RQ3, we consider the instances where at least one approach can solve within the time budget for computing this metric.
    \item Maximum verifiable perturbation ($\inpSpecDist^{*}$): the largest $\inpSpecDist$ for which global robustness can be certified, obtained via binary search over $\inpSpecDist$. This metric evaluates the effectiveness of our approach for handling problems in practice.
\end{compactitem}

\myparagraph{{Benchmarks}}
\begin{table}
    \centering
    \caption{Overview of benchmarks and network architectures. \# Neu.: the number of neurons, \# Ins.: the number of instances, $t_{\text{budget}}$: time budget. Time budgets are used in the experiment for RQ1, RQ2, and RQ3.}
    \label{tab:benchmarks}
    \setlength{\tabcolsep}{3pt} 
    \begin{tabular}{llcccc}
    \toprule
    \textbf{Benchmark} & \textbf{Architecture} & \textbf{\# Neu.} & \textbf{\# Ins.} & \textbf{$t_{\text{budget}}$}(s) \\
    \midrule
    \acasxu      & FNN (7-layer)          & 305   & 230 & 420 \\ 
    \mnistF       & FNN (5-layer)            & 1,034 & 156 & 600 \\
    \mnistC       & CNN (2 Cnv, 2 Lin)   & 9,518 & 156 & 600 \\
    \cifar     & CNN (2 Cnv, 2 Lin)   & 4,862 & 158 & 1800/3600/7200 \\
    \gtsrb   & CNN (2 Cnv, 2 Lin)  & 6,287 & 117 & 1800/3600/7200 \\
    \bottomrule
    \end{tabular}
\end{table}
Table~\ref{tab:benchmarks} presents the networks and corresponding instances used in our evaluation, where relational verification properties are formulated over standard datasets including \acasxu, \mnist, \cifarTen, and \gtsrb. Among these datasets, \acasxu has control-related background; \gtsrb involves image recognition in autonomous driving tasks; other benchmark sets, including \mnist and \cifar, consist of more general image classification tasks. These datasets are standard benchmarks adopted in VNN-COMP~\cite{brix2024fifth}, an annual competition of neural network verification.  

\begin{figure}[!tb]
    \centering
    \includegraphics[width=0.6\linewidth]{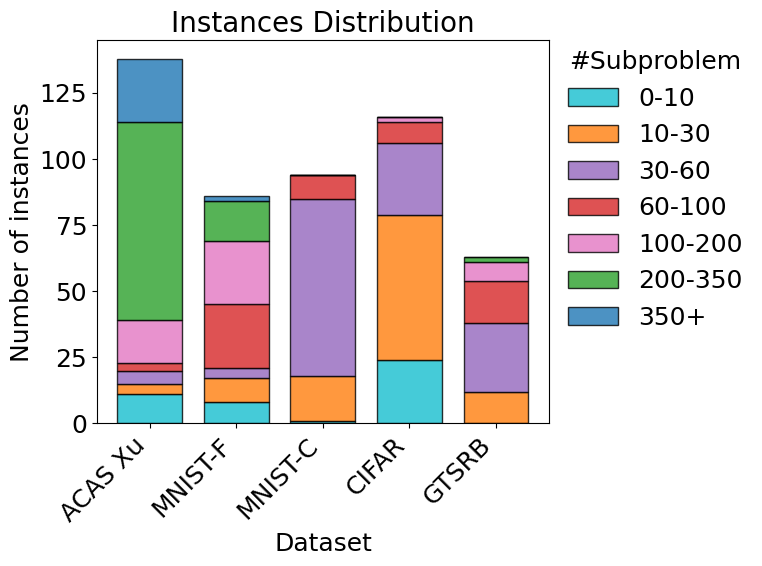}
    \caption{Instances distribution}
    \label{fig:instance_distr}
\end{figure}

While the original benchmarks typically target local robustness, our evaluation focuses on \emph{global robustness}, which requires certifying properties over pairs of inputs $(\inpOne, \inpTwo)$ over the same network within a bounded domain $\inpSpace$. To adapt the standard datasets for this relational setting, we formulate the verification instances by defining the domain $\inpSpace$ and the relational bound $\inpSpecDist$ as follows:

\begin{compactitem}
    \item {\acasxu}~\cite{katz2017reluplex}: $\inpSpace$ is derived directly from the operational envelopes defined in the original benchmark. We fix the relational distance to $\inpSpecDist = 0.1$.
    \item \mnist, \cifarTen, and \gtsrb: The input domain is defined as $\inpSpace = \{ x \mid \|x - x_{\mathrm{seed}}\|_\infty \leq i_{\mathrm{eps}}/256 \}$ and the relational distance as $\inpSpecDist = d_{\mathrm{eps}}/256$, where the denominator aligns with pixel values normalized to $[0,1]$. For \mnist and \cifarTen we evaluate nine $(i_{\mathrm{eps}}, d_{\mathrm{eps}}): \{(2,1),(3,1),(4,1),(4,2),(6,2),(8,2),(6,3),(9,3),\\(12,3)\}$ configurations, spanning a range of perturbation strengths from easy to hard. Since each sub-problem requires much longer time to solve as \inpSpecDist become larger in \cifar and \gtsrb, time limits  $(1800/3600/7200)$ are set on each $d_{\mathrm{eps}}$ $(1/2/3)$ individually.
\end{compactitem}
These settings cover a wide range of perturbations,  ensuring the rigor and fairness of our evaluation.
Fig.~\ref{fig:instance_distr} depicts the distribution of instances across each dataset. This indicates that the selected instances are broadly and evenly distributed, covering a wide and balanced range without bias.

\subsection{Evaluation Results}\label{sec:evaluationResults}


\researchquestion{Effectiveness of \tool~compared to approximation verifier \raven }\label{rq:raven}
\begin{table}
  \centering
\caption{RQ\ref{rq:raven}--Comparison with \raven}
  \label{tab:rq2}
   \resizebox{\linewidth}{!}{
  \begin{tabular}{lccccccccc}
    \toprule
    \multirow{2}{*}{Dataset} & 
    \multicolumn{2}{c}{\solved} & & 
    \multicolumn{2}{c}{\subproblems} & & 
    \multicolumn{2}{c}{\deltaTime (\%)} \\
    \cmidrule(lr){2-3} \cmidrule(lr){5-6} \cmidrule(lr){8-9}
     & \raven & \tool & & \raven & \tool & & \raven & \tool \\
    \midrule
    \acasxu & 42 & 67 && 1 & 116.3 && 0.79 & 20.50 \\
    \mnistF & 31 & 54 && 1 & 58.6 && 0.92 & 23.04 \\
    \mnistC & 23 & 27 && 1 & 25.9 && 9.56 & 66.07 \\
    \cifar & 10 & 23 && 1 & 23.5 && 2.40 & 46.79 \\
     \gtsrb &  9 &  33 &&  1 &  38.8 &&  2.22 &  39.09 \\
    \bottomrule
  \end{tabular}
   }
\end{table}

Table~\ref{tab:rq2} shows the overall performance comparison between \raven and \tool on various benchmarks and models.
In our experiment, we first apply \raven and then \tool if \raven fails to verify the property. Since \tool is equivalent to \raven if a problem can be solved by \raven  (i.e., the original problem at the root of a \bab tree), the number of solved instances of \tool is counted only for those that \raven could not solve.

From Table~\ref{tab:rq2}, we can observe that \tool significantly improves the verification performance compared to \raven in terms of the number of solved instances. While \tool generates more sub-problems and requires additional runtime due to iterative splitting, this additional computational cost is justified by its substantial improvement in the number of successfully verified instances.
This result indicates that \tool effectively enhances the verification capability for global robustness properties by leveraging relational branching strategies.

\researchquestion{Neuron splitting: Relational vs. individual} \label{rq:ri}
\begin{table*}
\centering
\caption{ RQ\ref{rq:ri}--Comparison between relational neuron splitting and individual neuron splitting}\label{tab:rq2ri}
\setlength{\tabcolsep}{5pt}
\begin{tabular}{lccccccccccccccc}
\toprule
\multirow{2}{*}{Method} & \multicolumn{3}{c}{\acasxu} & \multicolumn{3}{c}{\mnistF} & \multicolumn{3}{c}{\mnistC} & \multicolumn{3}{c}{\cifar} & \multicolumn{3}{c}{ \gtsrb} \\
\cmidrule(lr){2-4} \cmidrule(lr){5-7} \cmidrule(lr){8-10} \cmidrule(lr){11-13} \cmidrule(lr){14-16} & \solved &  \subproblems & \deltaTime & \solved & \subproblems & \deltaTime & \solved &  \subproblems & \deltaTime & \solved & \subproblems & \deltaTime & \solved & \subproblems & \deltaTime \\
\midrule
    \baseline & 12 & 133.9 & 87.51 & 23 & 135.9 & 79.24 & 6 & 32.0 & 89.49 & 28 & 21.6 & 71.82 & 24 & 50.9 & 60.95 \\
    \is & 9 & 117.3 & 89.46 & 27 & 124.7 & 73.22 & 8 & 33.3 & 90.86 & \tbgreen 31 & 21.4 & 68.63 & 23 & 58.0 & 66.83 \\
    \tool & \tbgreen 67 & 83.4 & 34.24 & \tbgreen 54 & 59.9 & 30.73 & \tbgreen 27 & 27.5 & 71.37 & 23 & 28.3 & 75.03 & \tbgreen 33 & 39.6 & 50.98 \\
\bottomrule
\end{tabular}
\end{table*}

\begin{figure}
    \centering
\begin{subfigure}{0.48\linewidth}
  \includegraphics[width=\linewidth]{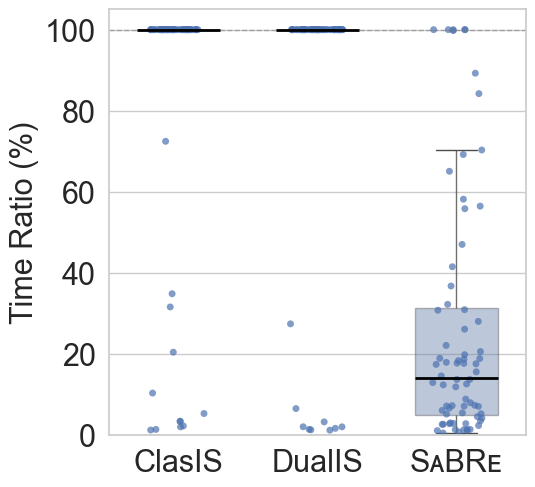}
  \caption{\acasxu}
  \label{fig:boxplot_aca}
\end{subfigure}
\hfill
\begin{subfigure}{0.48\linewidth}
  \includegraphics[width=\linewidth]{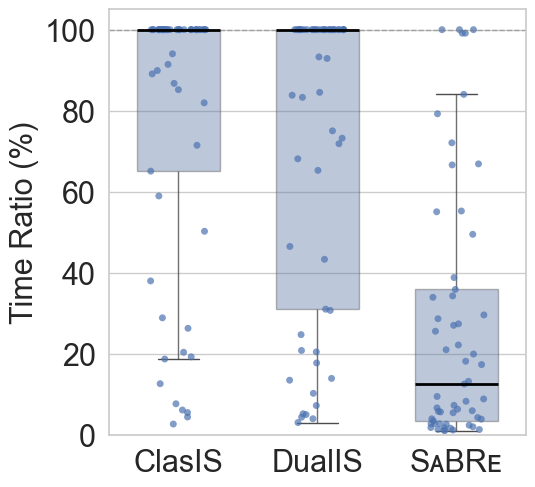}
  \caption{\mnistF}
  \label{fig:boxplot_mf}
\end{subfigure}
\\[0.5em]
\begin{subfigure}{0.48\linewidth}
  \includegraphics[width=\linewidth]{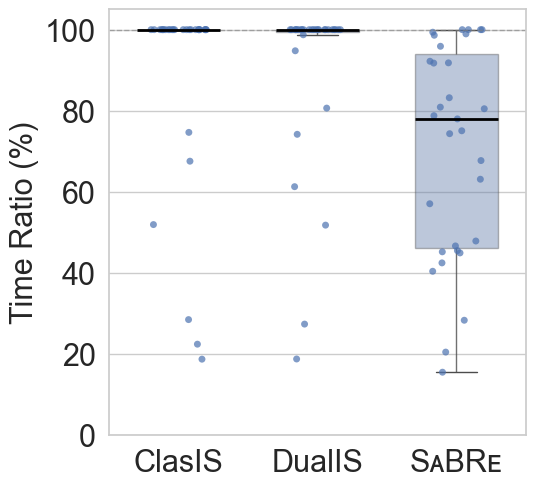}
  \caption{\mnistC}
  \label{fig:boxplot_mc}
\end{subfigure}
\hfill
\begin{subfigure}{0.48\linewidth}
  \includegraphics[width=\linewidth]{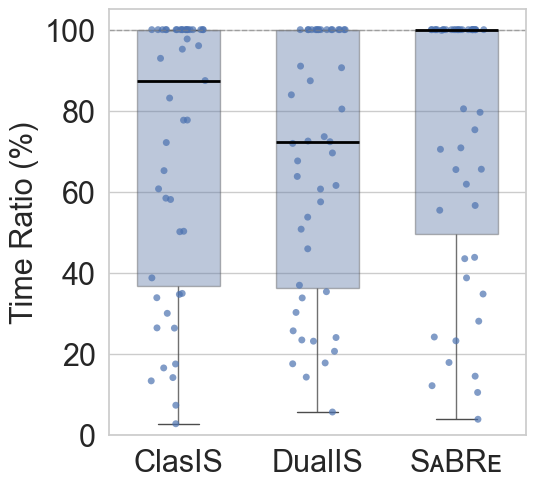}
  \caption{\cifar}
  \label{fig:boxplot_cif}
\end{subfigure}
\\[0.5em]
\begin{subfigure}{0.48\linewidth}
  \includegraphics[width=\linewidth]{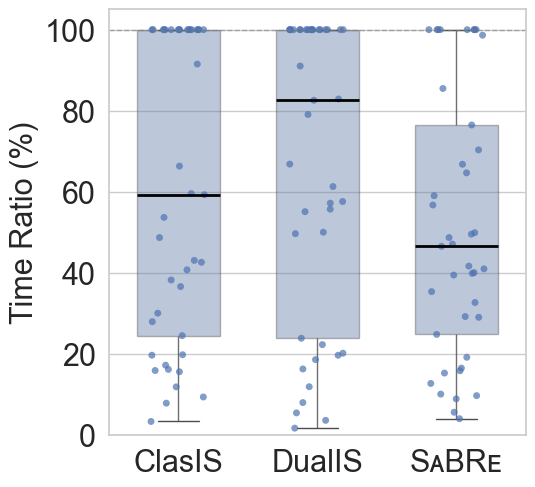}
  \caption{\gtsrb}
  \label{fig:boxplot_gtsrb}
\end{subfigure}
    \caption{RQ\ref{rq:ri}--Relational splitting (\tool) vs.\ individual splitting as verification progresses. 
    }
    \label{fig:rq2_box_all}
\end{figure}

Table~\ref{tab:rq2ri} compares the performance of relational neuron splitting (\tool) and individual neuron splitting (\baseline, \is) across five benchmarks. We observe that \tool consistently outperforms both \baseline and \is on \acasxu, \mnistF, \mnistC, and \gtsrb in terms of the number of solved instances (\solved), the number of sub-problems (\subproblems), and verification time ratio (\deltaTime). For fairness, the time ratio is computed only over instances for which at least one of the approaches successfully completes verification.

\begin{figure}[!tb]
  \centering
  \begin{subfigure}[b]{0.45\linewidth}
    \includegraphics[width=\linewidth]{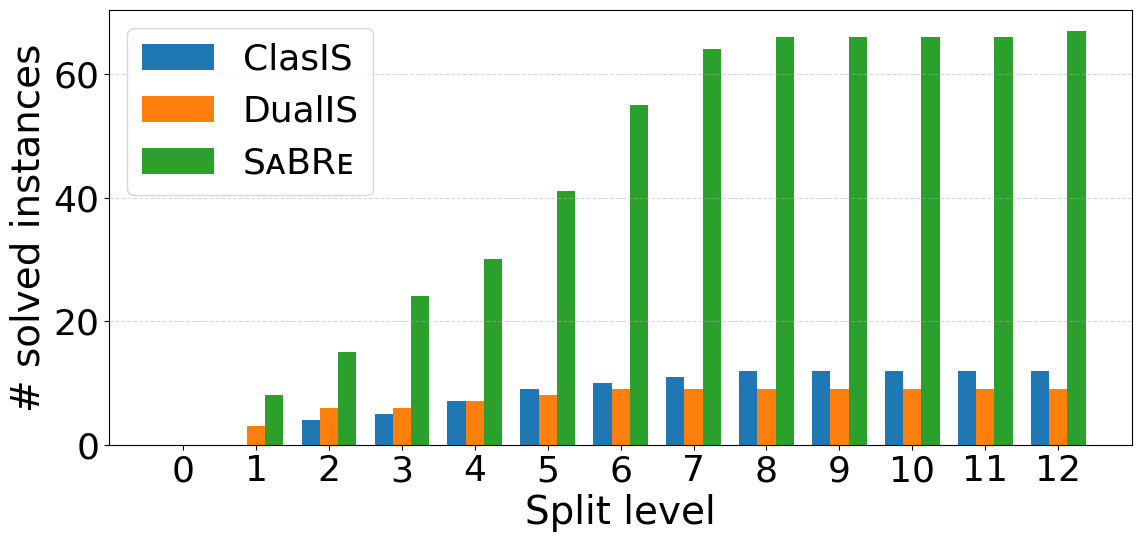}
    \caption{\acasxu}
    \label{fig:trans_aca}
  \end{subfigure}
  \hfill
  \begin{subfigure}[b]{0.45\linewidth}
    \includegraphics[width=\linewidth]{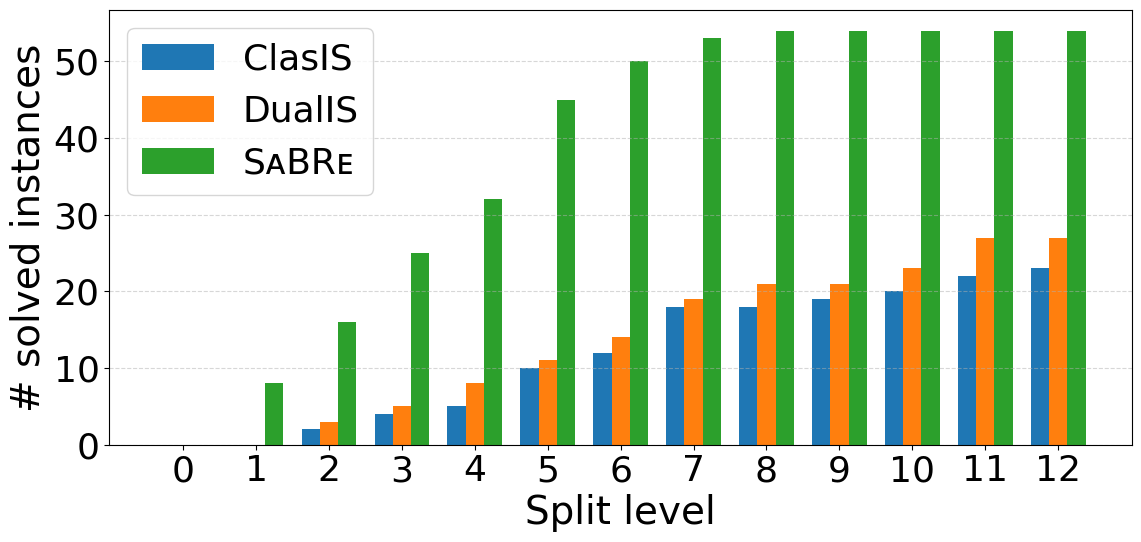}
    \caption{\mnistF}
    \label{fig:trans_mf}
  \end{subfigure}
  \\[0.5em]
  \begin{subfigure}[b]{0.45\linewidth}
    \includegraphics[width=\linewidth]{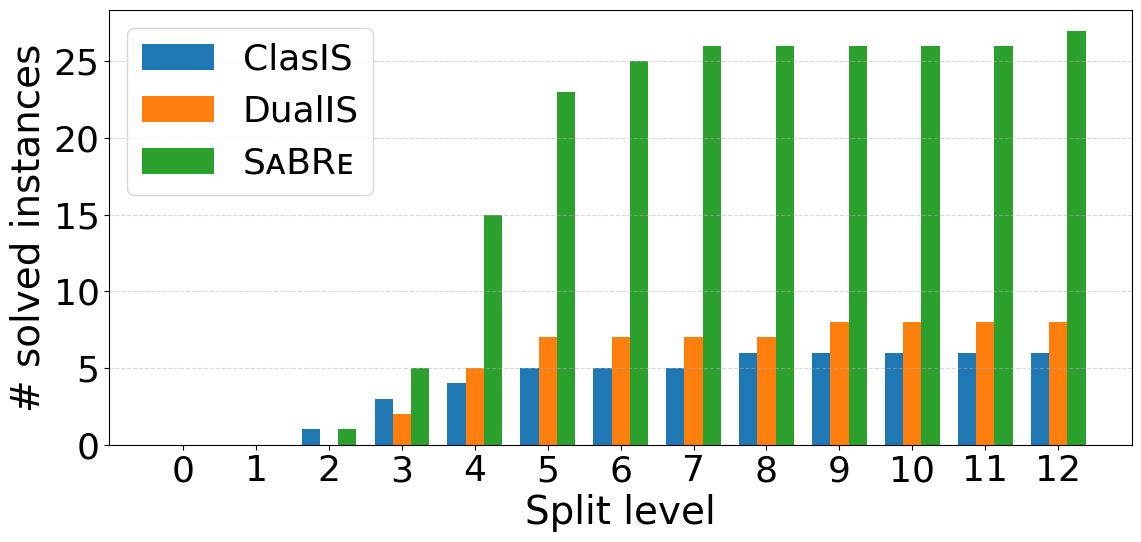}
    \caption{\mnistC}
    \label{fig:trans_mc}
  \end{subfigure}
  \hfill
  \begin{subfigure}[b]{0.45\linewidth}
    \includegraphics[width=\linewidth]{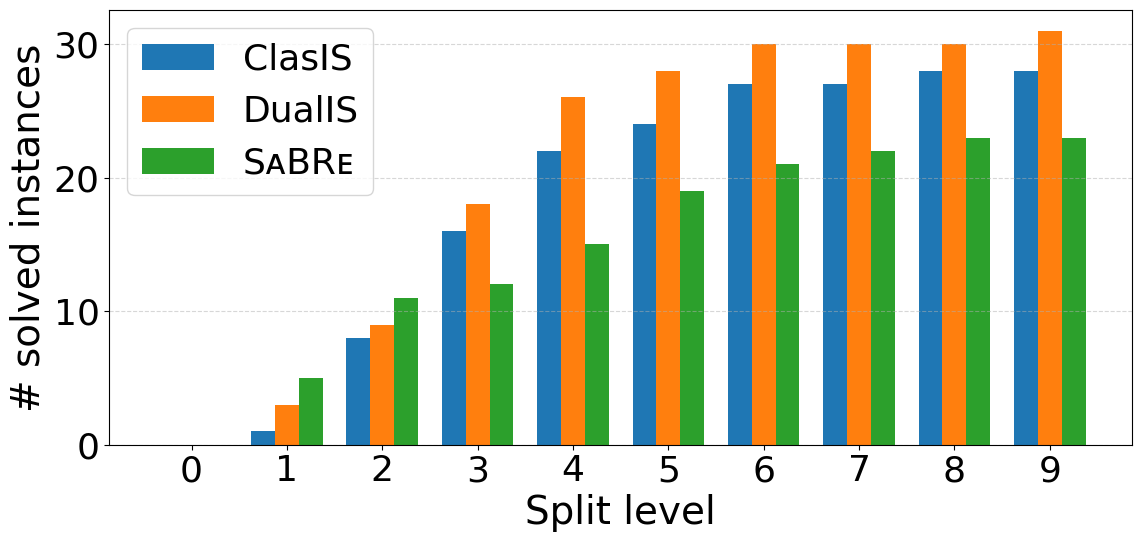}
    \caption{\cifar}
    \label{fig:trans_cif}
  \end{subfigure}
  \begin{subfigure}[b]{0.45\linewidth}
    \includegraphics[width=\linewidth]{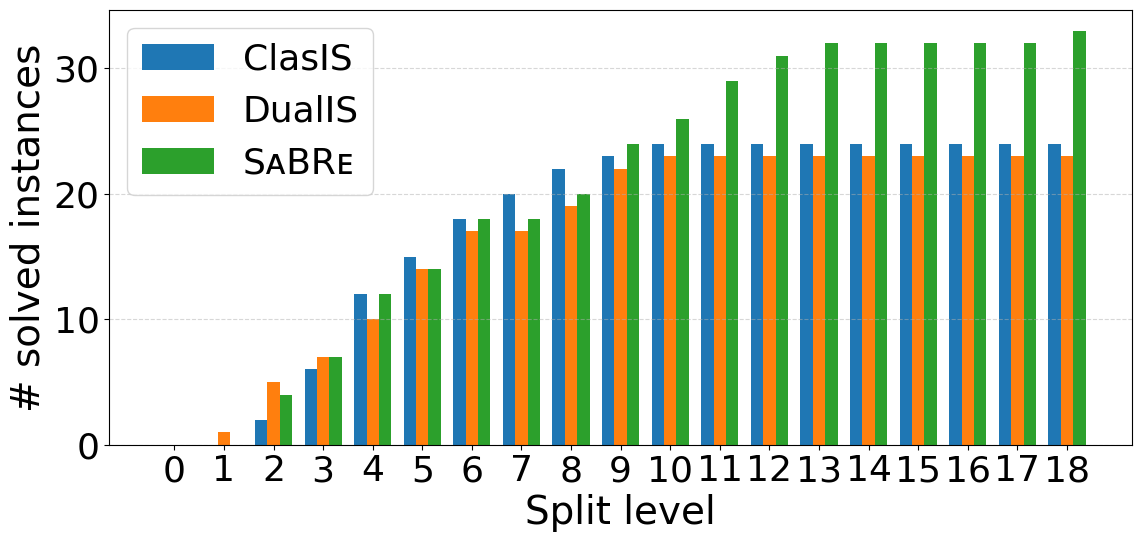}
    \caption{\gtsrb}
    \label{fig:trans_gtsrb}
  \end{subfigure}
  \caption{RQ\ref{rq:ri}--Cumulative number of verified instances against split level for each method.
  }
  \label{fig:rq2-trans}
\end{figure}

On \acasxu, \tool solves 67 instances while \is and \baseline solve only 9 and 12, 
respectively, completing verification tasks with more than $\approx\!30\%$ less number of sub-problems by effective splitting and achieving more than a 50\% reduction in average time ratio. 
On \mnistF, \tool solves roughly twice as many instances (54 vs.\ 27 and 23) while 
reducing the numbers of sub-problems by more than half in both cases and average time from $\approx\!76\%$ to $31\%$. 
A similar trend appears on \mnistC, where \tool verifies 27 instances compared to 
8 and 6 for \is and \baseline, with a smaller number of sub-problems (27.5 vs.\ 33.3 and 32.0) and a lower time ratio (71.37\% vs.\ 90.86\% and 89.49\%).
For \gtsrb, \tool solves 33 instances with a less number of sub-problems and shorter verification time, outperforming \is and \baseline, which solve only 23 and 24 instances, respectively.
However, on \cifar, individual splitting methods outperform \tool in both solved 
instances (31 and 28 vs.\ 23) and time efficiency (68.63\% and 71.82\% vs.\ 75.03\%).

Fig.~\ref{fig:rq2_box_all} confirms that these gains are consistent across the instance distribution: on \acasxu, both \mnist, and \gtsrb benchmarks, \tool's \deltaTime concentrates well below the timeout line with a notably lower median, while \baseline and \is accumulate many timeouts; on \cifar the pattern reverses, consistent with the numbers above. Fig.~\ref{fig:rq2-trans} provides a complementary perspective through cumulative split-level circumstances: on \acasxu, \mnistF, and \mnistC, \tool's curve rises more steeply and plateaus higher, indicating that each relational split resolves more instances than an individual split at equivalent depth. For \gtsrb, the three approaches perform comparably at small splitting levels. However, as the splitting level increases to $9$ and more, \tool outperforms the other approaches  and achieves the largest number of solved instances. On \cifar, \tool's curve rises more slowly, a behavior we analyze below. 

In summary, by relational neuron splitting, \tool can refine the approximation and verifies the problem more efficiently, compared to individual neuron splitting approaches. This is consistent with our expectation that, although our approach is not complete because of the strategy of splitting relational neurons, this selection also brings higher level of abstraction refinement and thus better performance, which is important to the adoption of verification in practice.

\myparagraph{Investigation for the performance difference between relational and individual splitting in \cifar}

\begin{table}
\centering
\caption{RQ\ref{rq:ri}--Relational neuron splitting vs. individual neuron splitting in \cifar} \label{tab:rq2_cifar}
\resizebox{\linewidth}{!}{
\begin{tabular}{lccccccccc}
\toprule
\multirow{2}{*}{Method} & \multicolumn{3}{c}{$\inpSpecDist=1/256$} & \multicolumn{3}{c}{$\inpSpecDist=2/256$} & \multicolumn{3}{c}{$\inpSpecDist=3/256$} \\
\cmidrule(lr){2-4} \cmidrule(lr){5-7} \cmidrule(lr){8-10} & \solved & \subproblems & \deltaTime & \solved & \subproblems & \deltaTime & \solved & \subproblems & \deltaTime \\
\midrule
\baseline & 16 & 22.1 & 56.49 & 8 & 22.6 & 77.34 & 4 & 19.3 & 93.54 \\
\is & 16 & 22.3 & 56.71 & 11 & 20.8 & 67.03 & 4 & 21.0 & 90.72 \\
\tool & 5 & 41.2 & 87.03 & 10 & 22.6 & 64.55 & 8 & 15.3 & 74.48 \\
\bottomrule
\end{tabular}
}
\end{table}

Table~\ref{tab:rq2_cifar} presents the comparison of relational and individual splitting under the different \inpSpecDist in \cifar. 

For smaller \inpSpecDist, particularly $\inpSpecDist = 1/256$, individual splitting methods (\baseline, \is) outperform relational splitting (\tool) in terms of the number of solved instances ($16,16$ vs. $5$) and time cost ($\approx\!56.5\%$ vs. $87.03\%$).
However, as \inpSpecDist increases, the performance of relational splitting gradually improves. On $\inpSpecDist = 2/256$, relational and individual splitting methods achieve comparable performance, while on $\inpSpecDist = 3/256$, relational splitting outperforms individual splitting in both metrics. In contrast to $4$ instances where individual splitting approaches achieve, \tool using relational splitting achieves $8$ instances within the shorter time cost.

These results suggest that relational splitting is less effective when the relational bounds are small, and becomes more effective as the propagated relational bounds increase. This matches the fact that the relational bounds are obtained by propagating input bounds through the network, and thus larger input relational distances produce wider bounds and make relational splitting more effective.

To understand why this performance transition under \inpSpecDist happens only in \cifar, we further analyze the model characteristics. Although \mnistC, \cifar, and \gtsrb models share the similar depth and layer structures, we observe that the \cifar model contains significantly more unstable \relu neurons (i.e., $\lowerBound{\indNeuronPre{i}{j}} < 0 < \upperBound{\indNeuronPre{i}{j}}$) along the reasoning path. These unstable neurons can enlarge the search space and reduce the effectiveness of relational splitting, as each split yields only a small local improvement. Consequently, the refinement from relational splits on the small \inpSpecDist becomes more global and diffuse.

\researchquestion{How does the performance of \tool scale under problems of different complexities?  }\label{rq:scalability_analysis}
\begin{table}[!tb]
\centering
\caption{RQ\ref{rq:scalability_analysis}--Performance comparison on different number of dimensional input perturbation. \perturbRate indicates the ratio of the number of input dimensions to add perturbation.}
\label{tab:scalability_analysis_result_summary}
\resizebox{\linewidth}{!}{
\begin{tabular}{lcccccccccccc}
\toprule
\multirow{2}{*}{Method} &
\multicolumn{3}{c}{$\perturbRate=0.125$} &
\multicolumn{3}{c}{$\perturbRate=0.25$} &
\multicolumn{3}{c}{$\perturbRate=0.5$} &
\multicolumn{3}{c}{$\perturbRate=1.0$} \\
\cmidrule(lr){2-4} \cmidrule(lr){5-7} \cmidrule(lr){8-10} \cmidrule(lr){11-13}
& \solved & \subproblems & \deltaTime
& \solved & \subproblems & \deltaTime
& \solved & \subproblems & \deltaTime
& \solved & \subproblems & \deltaTime \\
\midrule
\baseline & 39 & 31.6 & 9.56 & 28 & 125.6 & 37.71 & 20 & 164.8 & 52.62 & 15 & 144.4 & 58.72 \\
\is       & 40 & 31.6 & 9.24 & 28 & 121.6 & 34.91 & 18 & 152.5 & 53.28 & 17 & 113.4 & 49.73 \\
\tool     & 41 & 16.6 & 6.89 & 40 & 38.7 & 11.02 & 35 & 78.0 & 27.00 & 24 & 75.7 & 30.01 \\
\bottomrule
\end{tabular}
}
\end{table}

\begin{table}[!tb]
\centering
\caption{RQ\ref{rq:scalability_analysis}--The number of unstable \relus in two networks on different \perturbRate. Each cell represents ``mean \# unstable \relus'' \;/ ``\# total \relus''.}
\label{tab:the_number_of_unstable_relus}
\resizebox{\linewidth}{!}
{\begin{tabular}{lcccc}
\toprule
& $\perturbRate=0.125$ & $\perturbRate=0.25$ & $\perturbRate=0.5$ & $\perturbRate=1.0$ \\
\midrule
Unstable \relus & 17.2 / 2068 & 36.0 / 2068 & 83.4 / 2068 & 406.2 / 2068 \\
\bottomrule
\end{tabular}
}
\end{table}

In this RQ, we investigate how the performances of individual and relational splitting are affected by different factors relevant to problem complexities. 

Specifically, we consider the number of perturbed input dimensions as a controllable factor, and vary the number of perturbed input dimensions. We represent different numbers of perturbed input dimensions by \perturbRate, i.e., the product of $\perturbRate$ and the total number of dimensions. For each \perturbRate, we use 45 instances randomly selected from our benchmark set and introduce perturbations to a varying number of input dimensions. The perturbed dimensions are selected randomly.

Table~\ref{tab:scalability_analysis_result_summary} presents the performance comparison over different \perturbRate.
Overall, while the performance of \tool is comparable to other approaches when $\perturbRate = 0.125$ (especially for \solved), 
it is evidently superior to \is and \baseline when $\perturbRate$ increases to a higher level from $0.25$ to $1.0$, in terms of all metrics (\solved, \subproblems, and \deltaTime). 
For $\perturbRate = 0.125$, although different approaches solve similar numbers of problems, but \tool performs with lower cost (16.6 of \subproblems and 6.89 of \deltaTime) compared to \is (31.6 of \subproblems and 9.24 of \deltaTime) and \baseline (31.6 of \subproblems and 9.56 of \deltaTime). 
For $\perturbRate = 0.25$, compared to 28 instances solved by individual approaches (\is and \baseline), \tool solved 40 instances. Moreover, \tool also takes less runtime cost, with 38.7 of \subproblems and 11.02 of \deltaTime.
For $\perturbRate = 0.5$, \tool outperforms by solving 35 instances with average 78 sub-problems and 27 seconds. In contrast, \is and \baseline respectively solve 18 and 20 with around double runtime cost in terms of both \subproblems and \deltaTime.
For $\perturbRate = 1.0$, \tool records $24$ solved instances against $17$ and $15$ of \is and \baseline, respectively. For runtime cost, \tool takes less effort by $\approx33\%$ \subproblems and $\approx40\%$ \deltaTime. These results demonstrate that, \tool exhibits consistent and evident performance advantages over the baseline approaches across different numbers of input dimensions, which strengthens the effectiveness of our strategies.

Moreover, in Table~\ref{tab:the_number_of_unstable_relus} we also report the numbers of individual and relational unstable \relus under the approximation we adopted. As an additional indicator of problem complexity, although we cannot directly control it, we can observe that by increasing the number of input dimensions, the ratio of unstable \relus also increase. Therefore, our observations regarding the performance changes with respect to unstable \relus ratio, is similar to that with numbers of input dimensions, namely, as ratio of unstable \relus increases, our performance advantages are consistent and evident over other approaches.

\researchquestion{Performance comparison of neuron selection}\label{rq:ns}
\begin{table*}[!t]
  \centering
\caption{RQ\ref{rq:ns}--Comparison of neuron selection \tool \emph{v.s.} \RSrandom.}
  \label{tab:rq4-comparison_dual_random}
  \begin{tabular}{lccccccccccccccc}
    \toprule
    \multirow{2}{*}{Method} & \multicolumn{3}{c}{\acasxu} & \multicolumn{3}{c}{\mnistF} & \multicolumn{3}{c}{\mnistC} & \multicolumn{3}{c}{\cifar} & \multicolumn{3}{c}{\gtsrb} \\
    \cmidrule(lr){2-4} \cmidrule(lr){5-7} \cmidrule(lr){8-10} \cmidrule(lr){11-13} \cmidrule(lr){14-16} & \solved & \subproblems & \deltaTime & \solved & \subproblems & \deltaTime & \solved & \subproblems & \deltaTime & \solved & \subproblems & \deltaTime & \solved & \subproblems & \deltaTime \\
    \midrule
    \RSrandom
    & 44 & 199.2 & 60.35
    & 44 & 117.6 & 48.11
    & 11 & 38.7 & 90.43
    & 8 & 24.4 & 87.17
    & 9 & 92.5 & 86.60 \\
    
    \tool
    & \tbgreen 67 & 100.8 & 34.24
    & \tbgreen 54 & 67.3 & 30.73
    & \tbgreen 27 & 28.2 & 71.37
    & \tbgreen 23 & 18.0 & 75.03
    & \tbgreen 33 & 35.5 & 39.09 \\
	\bottomrule
  \end{tabular}
\end{table*}

Table~\ref{tab:rq4-comparison_dual_random} evaluates the performance of relational neuron selection strategies by comparing our proposed tool \tool and the \RSrandom baseline. 

The results show that our selection strategy leads to substantial improvements across all benchmarks compared to random selection.

On \acasxu, our method solves 67 instances, compared to only 44 solved by random selection, and reduces the sub-problem exploration cost by around half and the verification time from 60.35\% to 34.24\%.
On \mnistF, the number of solved instances increases from 44 to 54, while the number of sub-problems and the average time drops from 117.6 to 67.3 and from 48.11\% to 30.73\%, respectively.
On \mnistC, our method solves 27 instances, more than double of the 11 solved by random selection, while achieving smaller exploration cost for sub-problems (28.2 vs. 38.7) and a lower time ratio (71.37\% vs. 90.43\%).
On \cifar, our strategy solves 23 instances compared to only 8 under random selection, and reduces the number of sub-problems from 24.4 to 18 and the verification time from 87.17\% to 75.03\%.
Finally on \gtsrb, \tool achieves 33, which is around 4 times compared to 9 of \RSrandom. Similar to others, the number of sub-problems and runtime also decrease by $\approx 50$ and 50\%, respectively.

These results show our selection strategy leads to substantial improvements across all benchmarks compared to random selection, demonstrating that selection strategy plays an important role in relational \bab.

\researchquestion{How does \tool compare in terms of maximum verifiable perturbation?}\label{rq:max}

To demonstrate the usefulness of our approach, we conduct a binary search over $\inpSpecDist$ to obtain the maximum perturbation $\inpSpecDist^{*}$ for each instance and method. The initial search intervals are $[0,\, 1]$ for \acasxu, $[0,\, 12/256]$ for \mnist, $[0,\, 8/256]$ for \cifar, and $[0,\, 12/256]$ for \gtsrb, with a termination tolerance of $10^{-6}$. For \mnist and \cifar, a minimum step size of $1/256$ is enforced to respect pixel-level clamp. 
We set the time budget for both: a verification step for a single \inpSpecDist and entire process of binary search. The pairs of time $(\textit{single step}(s), \textit{entire process}(s))$ for \acasxu, \mnistF, \mnistC, \cifar, and \gtsrb are $(420, 1200)$, $(600, 2700)$, $(800, 2700)$, $(4000, 18000)$, and $(4000, 18000)$, respectively.

\begin{figure*}[!t]
    \centering
    \includegraphics[width=\textwidth]{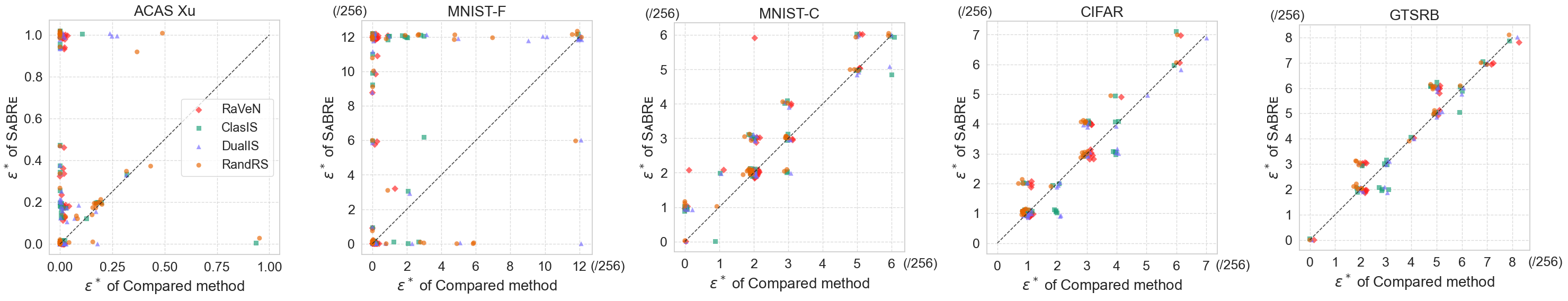}
    \caption{RQ\ref{rq:max}--Maximum verifiable relational input distance ($\inpSpecDist^{*}$ obtained by binary search. Each point is one instance, where the $x$-axis is the compared method's $\inpSpecDist^{*}$ and
    the $y$-axis is \tool's. Points above the diagonal indicate \tool certifies a larger perturbation region.}
    \label{fig:bs_diag}
\end{figure*}

\begin{figure}[!tb]
    \centering
    \includegraphics[width=0.7\linewidth]{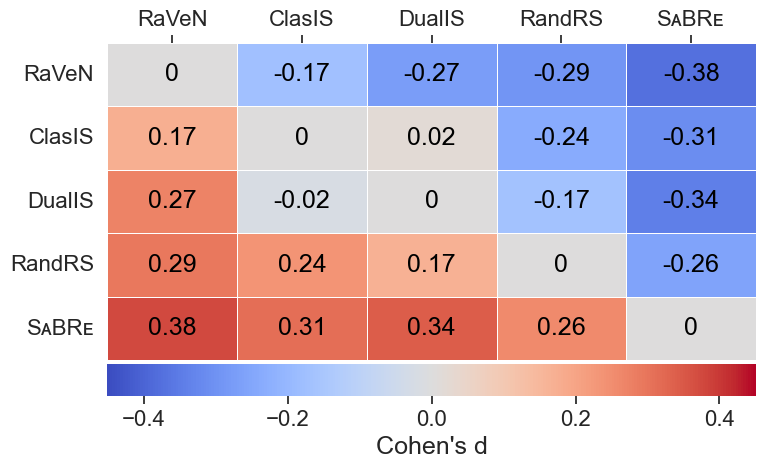}
    \caption{RQ\ref{rq:max}--Pairwise effect size (Cohen's $d$) aggregated across all datasets. Positive values favour the row method.}
    \label{fig:statistical_comparison}

\end{figure}

\tool consistently achieves the largest certifiable perturbation region across the majority of instances, with the advantage being largest on \acasxu and \mnistF with the same time budget. As shown in Fig.~\ref{fig:bs_diag}, on these two benchmarks a substantial fraction of points lie on the \emph{vertical axis}, where all competing methods return $\inpSpecDist^{*} = 0$ while \tool produces non-zero bounds. This reflects a qualitative advantage: other methods fail to certify \emph{any} non-trivial perturbation within the time budget, whereas \tool still produces meaningful bounds and manages to complete the relational verification task. On \mnistC,
the dominance of \tool is more uniform, with most points lying above the diagonal line. The \cifar plot reveals a more nuanced picture: \tool dominates \raven or \RSrandom and records equivalent performance overall against \is. For \baseline, \tool reaches larger $\inpSpecDist^{*}=5/256$ and $\inpSpecDist^{*}=7/256$, but in total, achieves one less instances with larger $\inpSpecDist^{*}$. 
On \gtsrb, \tool consistently outperforms \raven and \RSrandom. Compared to \is, \tool manages to find larger $\inpSpecDist^{*}$ in 5 instances, while it loses in 2 instances.  
Compared to \baseline, \tool shows similar results in terms of the number of instances with higher values, specifically, in 5 instances 
\tool manages to find larger $\inpSpecDist^{*}$ while in 5 instances \tool fails. While we observe the instances where \baseline achieves higher certifiable value, i.e., the instances whose $\inpSpecDist^{*}=3/256$ by \baseline, 
we also observe the instance where \tool wins, such as the instances whose $\inpSpecDist^{*}=6/256$ by \tool.

To compare different approaches more rigorously, we apply statistical analysis to analyze the results. As shown in Fig.~\ref{fig:statistical_comparison}, \tool holds a consistent advantage over all baselines, with moderate effect sizes ranging from $d = 0.26$ to $d = 0.38$. Here, each cell reports statistical Cohen's $d$~\cite{cohen2013statistical}
computed over the per-instance $\inpSpecDist^{*}$ values, where a positive value indicates that the row method achieves larger certifiable perturbation regions on average. The fact that these gains are stable across five benchmarks spanning different architectures and data domains suggests that the benefit of relational neuron splitting is not tied to any particular network type or perturbation regime.

\section{Related work}\label{sec:related}
\myparagraph{Verification against local robustness}
Neural network verification against local robustness has been extensively studied~\cite{tjeng2018evaluating,katz2017reluplex,ehlers2017formal,huang2017safety,singh2019abstract,muller2020neural,wang2018efficient, shi2022efficiently} and approximated methods are often preferable thanks to their efficiency~\cite{anderson1811strong,tjandraatmadja2020convex, singh2019beyond,muller2022prima,raghunathan2018semidefinite}.
While approximation methods contributes to the scalability, they may raise false alarms due to the completeness issue. To tackle this issue, studies~\cite{henriksen2021deepsplit,de2021improved,shi2024neural,luneural,ferrari2022complete,wang2021beta,bunel2020branch} strike a balance between efficiency and completeness by incorporating with \bab as discussed in \S{\ref{sec:bab_local}}. 

\myparagraph{Relational verification}
This is an emerging topic and the related research still remains at an early stage. Nevertheless, existing research has investigated both exact methods~\cite{xie2023deepgemini} and approximation methods~\cite{wang22itne,banerjee24raven,banerjee24racoon}. As exact methods are exhaustive and suffer from scalable issues, we consider approximation methods in our evaluation. As discussed in \S{\ref{sec:OverAppVerification}}, \raven provides a systematic approximate bounding approach called \diffpoly. It enables the tighter linear relaxation based on the pre-activation states of individual and relational neurons. However, analogous to the verification against local robustness, approximation approaches including \raven suffer from the completeness issue. To address it, \rabbit~\cite{suresh24rabbit} integrates \bab into the relational verification, but its strategy of selecting and splitting individual neurons can bring suboptimal performance when applied in our settings (global robustness), as we show in our evaluation in~\S{}\ref{sec:experiment}. 
In comparison to individual neuron splitting, we explore the strategy of relational splitting that aims to directly split relational neurons and effectively reduce the over-approximation on the relational bounds.

\myparagraph{Differential/Incremental verification} Several verification settings involve reasoning about multiple network executions but differ from relational 
verification in key aspects. \emph{Differential verification}~\cite{paulsen2020neurodiff,paulsen2020reludiff,teuber2025revisiting} certifies two distinct networks produce similar outputs on the same input. \emph{Incremental verification}~\cite{fischer2022shared,ugare2022proof,ugare2023incremental,zhang2025efficient} reuses intermediate results from verifying an original network to accelerate verification of its updated (i.e., fine-tuned) variant. In contrast, relational verification considers two identical networks and certifies output consistency across \emph{different} inputs, and our work refines this overapproximation reasoning directly via relational neuron splitting.

\section{Conclusion and Future Work}\label{sec:conclusion}
We presented \tool, a \bab framework for relational neural network verification that splits relational neurons rather than individual neurons,
paired with a neuron selection strategy based on dual problem formulation of the relational neuron propagation. Our evaluation on 817 instances across \acasxu, \mnist, \cifar, and \gtsrb demonstrates that relational neuron splitting consistently outperforms individual neuron splitting, and that the dual-based selection strategy provides reliable guidance for prioritizing which relational neuron to split at each step. Together, these designs enable \tool to certify larger perturbation regions and resolve more verification instances within the time budget than existing approaches on most benchmarks. 

As future work, we plan to investigate more refined splitting strategies to achieve tighter relational bounds and design neuron selection heuristic to further improve the efficiency and scalability of verification. Moreover, for problem splitting,  our current approach selects relational neurons exclusively without considering individual neurons. While it is possible to consider both individual and relational neurons simultaneously, it introduces a non-trivial question about when we should select individual neurons and when we switch back to relational neurons. A possible solution could be based on our dual formulation, in which both individual and relational neurons are involved, however, more algorithmic details require more sophisticated design and comprehensive evaluation.

\bibliographystyle{IEEEtran}
\bibliography{mybib}

\appendices

\section{Dual Formulation for ReLU selection in classic \bab}\label{appendix:reluSelectionClassic}
Before presenting the dual formulation for individual networks, we first summarize the symbolic propagation for linear and ReLU layers for individual networks.

{\mathcompact
\begin{equation}
\begin{aligned}
\textbf{Linear} \\
&x\uid{i+1} = W\uid{i}y\uid{i} + b\uid{i}, \; \forall i \in \{ 0, ..., \outLayerIdx-1 \}\\
\textbf{ReLU} \\
&\indNeuron{i}{j} = 0, \; \forall i \in \{ 1, ..., \outLayerIdx-1 \}, \; j \in \reluNeg_i\\
&\indNeuron{i}{j} = \indNeuronPre{i}{j}, \; \forall i \in \{ 1, ..., \outLayerIdx-1 \}, \; j \in \reluPos_i\\
&\left.\begin{aligned}
  &\indNeuron{i}{j} \geq \slope{i}{j} \indNeuronPre{i}{j}\\
  &(\upperBound{\indNeuronPre{i}{j}} - \lowerBound{\indNeuronPre{i}{j}})\indNeuron{i}{j} - ( \upperBound{\indNeuronPre{i}{j}} \indNeuronPre{i}{j} - \upperBound{\indNeuronPre{i}{j}} \lowerBound{\indNeuronPre{i}{j}} ) \leq 0
\end{aligned}\right\} \; \begin{aligned}
    &\forall i \in \{ 1, ..., \outLayerIdx-1 \}, \; \\ &j \in \reluUns_i
\end{aligned}\\   
\end{aligned}
\end{equation}}
where $\slope{i}{j} \in [0, 1]$ handles the slope of lower bound of \relu~relaxation. $\reluNeg$, $\reluPos$, and $\reluUns$ are sets of indexes, representing the pre-activation states of $\indNeuronPre{i}{j}$ in layer $i$ where $\upperBound{\indNeuronPre{i}{j}} \leq 0$, $\lowerBound{\indNeuronPre{i}{j}} \ge 0$, or $\lowerBound{\indNeuronPre{i}{j}} < 0 < \upperBound{\indNeuronPre{i}{j}}$, respectively.

\paragraph{Dual Formulation for Individual Networks}
In local robustness verification, the output comparison is typically encoded by an objective vector $C$ for the output vector $\inpOne\uid{\outLayerIdx} \in \outShape$ and given two output dimensions $i$ and $j$:
\begin{equation}
\text{min} \quad \inpOne\uid{\outLayerIdx}_i - \inpOne\uid{\outLayerIdx}_j = C\inpOne\uid{\outLayerIdx}
\end{equation}
where $\inpOne\uid{\outLayerIdx}_i \in \R$ denotes $i$-th dimensional output of the network. This minimization of output difference is converted into maximization dual objective:

{\mathcompact
\begin{equation}\label{eq:dual_obj_individual}
\begin{aligned}
\text{max}\\
& -(\inpOne + \inpSpecDist)\transposed [\preReluInd[0]]_-
+ (\inpOne - \inpSpecDist)\transposed [\preReluInd[0]]_+ \\
& - \sum_{i=0}^{\outLayerIdx - 1} \affineInd[i+1] \transposed b\uid{i} + \sum_{i=1}^{\outLayerIdx - 1} \sum_{j} \frac{\upperBound{\indNeuronPre{i}{j}} \lowerBound{\indNeuronPre{i}{j}}}{\upperBound{\indNeuronPre{i}{j}} - \lowerBound{\indNeuronPre{i}{j}}} [\preReluInd[i]_j]_{+} \\
\end{aligned}
\end{equation}
where
\begin{equation}\label{eq:dual_const_individual}
\begin{aligned}
	& \affineInd[\outLayerIdx] = -C \\
	& \preReluInd[i] = W\uid{i}\transposed \affineInd[i + 1] \quad \forall i \in \{ \outLayerIdx-1, \ldots, 0 \}\\
	& \affineInd[i]_j = \left\{\quad\begin{aligned}
		& 0 &\quad\text{if}\quad \upperBound{\indNeuronPre{i}{j}} \leq 0 \\
		& \preReluInd[i]_j &\quad\text{if}\quad 0 \leq \lowerBound{\indNeuronPre{i}{j}} \\
		& \frac{\upperBound{\indNeuronPre{i}{j}}}{\upperBound{\indNeuronPre{i}{j}} - \lowerBound{\indNeuronPre{i}{j}}} [\preReluInd[i]_j]_{+} - \slope{i}{j} [\preReluInd[i]_j]_{-} &\quad\text{if}\quad \lowerBound{\indNeuronPre{i}{j}} < 0 < \upperBound{\indNeuronPre{i}{j}} \\
	\end{aligned}\right. \\
    & \quad \forall i \in \{ \outLayerIdx-1, \ldots, 1\} \\
\end{aligned}
\end{equation}}
Here, 
\begin{compactitem}[$\bullet$]
    \item $\affineInd[i],\preReluInd[i]$ are dual variables associated with the affine and activation constraints in the dual network,
    \item $[z]_- = \max(0,-z)$, $[z]_+ = \max(0,z)$
\end{compactitem}
This dual formulation can be derived by following the same flow in Appendix~\ref{appendix:rel_dual}.

\paragraph{Split Impact Estimation in \babsr}
For an unstable neuron $\inpOne\uid{i}_j$, \babsr~estimates how much the objective bound would improve if this neuron is split.

Let $\affineInd[i]_j$ denote the dual variable before splitting, and let $\affineIndPrime[i]_j$ denote the dual variable after imposing the split constraint on one branch (either active or inactive).
When a split is performed, the ReLU relaxation for $\inpOne\uid{i}_j$ is replaced by an exact constraint corresponding to the chosen branch.
As a result, the linear relaxation constraints of the \relu transformation are removed.
Accordingly, the dual variable $\affineInd[i]_j$ is updated by following the dual backpropagation rule (Eq.~\ref{eq:dual_const_individual}) under the selected branch.
More concretely, $\affineIndPrime[i]_j$ becomes either $0$ or $\preReluInd[i]_j$, depending on whether the neuron is fixed to the inactive or active, respectively.

Because the relaxation constraints are removed after splitting, the last term in the dual objective (Eq.~\ref{eq:dual_obj_individual}) corresponding to the \relu relaxation becomes zero.
{\mathcompact
\begin{equation}
\frac{\upperBound{\indNeuronPre{i}{j}} \lowerBound{\indNeuronPre{i}{j}}}{\upperBound{\indNeuronPre{i}{j}} - \lowerBound{\indNeuronPre{i}{j}}} [\preReluInd[i]_j]_{+} = 0
\end{equation}}

Therefore, the change in the dual objective can be estimated by substituting $\affineIndPrime[i]_j$ into the dual objective and measuring the resulting difference.
By considering the second line of the dual objective in Eq.~\ref{eq:dual_obj_individual}, the estimated improvement caused by splitting neuron $\inpOne\uid{i}_j$ can be expressed as:

{\mathcompact
\begin{equation}\label{eq:babsr_estimation}
  \left( \affineInd[i]_j - \affineIndPrime[i]_j \right) b\uid{i-1}_j + \frac{\upperBound{\indNeuronPre{i}{j}} \lowerBound{\indNeuronPre{i}{j}}}{\upperBound{\indNeuronPre{i}{j}} - \lowerBound{\indNeuronPre{i}{j}}} [\preReluInd[i]_j]_{+}
\end{equation}}

\section{\diffpoly: \relu~Transformation in Relational Setting}\label{appendix:relu_transformation_raven}
As described in \S{}~\ref{sec:OverAppVerification}, relational bounds are propagated symbolically layer by layer. 
Table~\ref{tab:diffpoly_1} and~\ref{tab:diffpoly_2} summarize the conditional approximate symbolic bound propagation rules at \relu implemented in \diffpoly (\raven~\cite{banerjee24raven}). In both tables, the first column shows the conditional case, the second column shows its symbolic lower and upper bound. We denote them $\relNeuronPost{i}{j,L}$ and $\relNeuronPost{i}{j,U}$, respectively. $\reluStateOne$, $\reluStateTwo$, and $\reluStateRel$ are the sets of indices $j$ grouped according to the pre-activation states of $\indNeuronPre{i}{j}$, $\indNeuronTwoPre{i}{j}$, and $\relNeuronPre{i}{j}$ (e.g., $j \in \reluUnsHat_i$ if $\lowerBound{\indNeuronTwoPre{i}{j}} < 0 < \upperBound{\indNeuronTwoPre{i}{j}}$).

The detailed case analysis over the pre-activation states of $\inpOne$, $\inpTwo$, and the relational neuron $\relNeuron\inpOne$ enables a more precise approximation of the nonlinear \relu~transformation and significantly reduces relaxation looseness during bounds propagation. 

\begin{table}[H]
\centering
\renewcommand{\arraystretch}{1.4}
\begin{tabular}{c|c}
\toprule
\textbf{Case} & \textbf{Symbolic bounds for $\relNeuronPost{i}{j}$} \\
\midrule

$j \in \reluNeg_i \cap \reluNegHat_i$ &
$\relNeuronPost{i}{j,L} = 0,\;
 \relNeuronPost{i}{j,U} = 0$ \\

$j \in \reluPos_i \cap \reluNegHat_i$ &
$\relNeuronPost{i}{j,L} = \indNeuronPre{i}{j},\;
 \relNeuronPost{i}{j,U} = \indNeuronPre{i}{j}$ \\

$j \in \reluNeg_i \cap \reluPosHat_i$ &
$\relNeuronPost{i}{j,L} = -\indNeuronTwoPre{i}{j},\;
 \relNeuronPost{i}{j,U} = -\indNeuronTwoPre{i}{j}$ \\

$j \in \reluPos_i \cap \reluPosHat_i$ &
$\relNeuronPost{i}{j,L} = \relNeuronPre{i}{j},\;
 \relNeuronPost{i}{j,U} = \relNeuronPre{i}{j}$ \\

$j \in \reluUns_i \cap \reluNegHat_i$ &
$\relNeuronPost{i}{j,L} = y_L,\;
 \relNeuronPost{i}{j,U} = y_U$ \\

$j \in \reluNeg_i \cap \reluUnsHat_i$ &
$\relNeuronPost{i}{j,L} = -\hat{y}_U,\;
 \relNeuronPost{i}{j,U} = -\hat{y}_L$ \\

$j \in \reluUns_i \cap \reluPosHat_i$ &
$\relNeuronPost{i}{j,L} = y_L - \indNeuronTwoPre{i}{j},\;
 \relNeuronPost{i}{j,U} = y_U - \indNeuronTwoPre{i}{j}$ \\

$j \in \reluPos_i \cap \reluUnsHat_i$ &
$\relNeuronPost{i}{j,L} = \indNeuronPre{i}{j} - \hat{y}_U,\;
 \relNeuronPost{i}{j,U} = \indNeuronPre{i}{j} - \hat{y}_L$ \\

$j \in \reluUns_i \cap \reluUnsHat_i$ &
$\relNeuronPost{i}{j,L} = y_L - \hat{y}_U,\;
 \relNeuronPost{i}{j,U} = y_U - \hat{y}_L$ \\
\bottomrule
\end{tabular}
\caption{Systematic \relu~symbolic propagation of relational neuron $\relNeuronPost{i}{j}$ based on the states of $\inpOne$ and $\inpTwo$}
\label{tab:diffpoly_1}
\end{table}

\begin{table}[htbp]
\centering
\renewcommand{\arraystretch}{1}
\begin{tabular}{c|c}
\toprule
\textbf{Case} & \textbf{Symbolic bounds for $\relNeuronPost{i}{j}$} \\
\midrule
$j \in \relReluPos_i$ & 
$
\relNeuronPost{i}{j,L} = 0,
\relNeuronPost{i}{j,U} = \relNeuronPre{i}{j}
$ \\
$j \in \relReluNeg_i$ & 
$
\relNeuronPost{i}{j,L} = \relNeuronPre{i}{j},
\relNeuronPost{i}{j,U} = 0
$ \\
$j \in \relReluUns_i$ & 
$\begin{aligned}
&\relNeuronPost{i}{j,L} = \frac{-\lowerBound{\relNeuronPre{i}{j}}(\relNeuronPre{i}{j} - \upperBound{\relNeuronPre{i}{j}})}{\upperBound{\relNeuronPre{i}{j}} - \lowerBound{\relNeuronPre{i}{j}}},\\
&\relNeuronPost{i}{j,U} = \frac{\upperBound{\relNeuronPre{i}{j}}(\relNeuronPre{i}{j} - \lowerBound{\relNeuronPre{i}{j}})}{\upperBound{\relNeuronPre{i}{j}} - \lowerBound{\relNeuronPre{i}{j}}}
\end{aligned}
$ \\
\bottomrule
\end{tabular}
\caption{Systematic \relu~symbolic propagation of relational neuron $\relNeuronPost{i}{j}$ based on the states of $\relNeuronPre{i}{j}$}
\label{tab:diffpoly_2}
\end{table}
These propagation rules form the basis for the dual formulation in the relational setting, which we introduce in the following appendix~\ref{appendix:rel_dual}.


\section{Dual Formulation in Relational Setting}\label{appendix:rel_dual}
In this section we derive Dual formulation for relational setting in \S{~\ref{sec:rel_neuron_selection}}.
We proceed the derivation on one representative \relu case: \exampleCase, but all other cases follow it analogously.

\subsection{Primal problem}\label{sec:primal_prob}
The relational verification objective can be expressed as the minimization of the output relational distance between two networks by taking a one-hot vector C, such as:
{\mathcompact
\begin{equation}\label{eq:primal_objective}
\begin{aligned}
    \text{min} \quad \relNeuronPost{\outLayerIdx}{\labelIdx} 
    &= \indNeuron{\outLayerIdx}{\labelIdx} - \indNeuronTwoPost{\outLayerIdx}{\labelIdx} \\
    &= C \cdot \relNeuronPost{\outLayerIdx}{}
\end{aligned}
\end{equation}}
The constraints corresponding to inputs, affine transformations and \relu transformation for individual neurons are
{\mathcompact
\begin{equation}\label{eq:primal_constraints}
\begin{aligned}
& \indNeuronOne{0}{} \geq \inpSpaceLower{}, \; \indNeuronOne{0}{} \leq \inpSpaceUpper{} \\
& \indNeuronTwo{0}{} \geq \inpSpaceLower{}, \; \indNeuronTwo{0}{} \leq \inpSpaceUpper{} \\
& \relNeuronPost{0}{} \geq -\inpSpecDist, \; \relNeuronPost{0}{} \leq \inpSpecDist\\
&\left.\begin{aligned}
  &x\uid{i+1} = W\uid{i}y\uid{i} + b\uid{i}\\
  &\hat{x}\uid{i+1} = W\uid{i}\hat{y}\uid{i} + b\uid{i}\\
  &\relNeuronPre{i+1}{} = W\uid{i}\relNeuronPost{i}{}\\
\end{aligned}\right\}\; \forall i \in \{ 0, ..., \outLayerIdx-1 \}\\
&\indNeuron{i}{j} = 0, \; \forall i \in \{ 1, ..., \outLayerIdx-1 \}, \; j \in \reluNeg_i\\
&\indNeuron{i}{j} = \indNeuronPre{i}{j}, \; \forall i \in \{ 1, ..., \outLayerIdx-1 \}, \; j \in \reluPos_i\\
&\indNeuronTwoPost{i}{j} = 0, \; \forall i \in \{ 1, ..., \outLayerIdx-1 \}, \; j \in \reluNegHat_i\\
&\indNeuronTwoPost{i}{j} = \indNeuronTwoPre{i}{j}, \; \forall i \in \{ 1, ..., \outLayerIdx-1 \}, \; j \in \reluPosHat_i\\
&\left.\begin{aligned}
  &\indNeuron{i}{j} \geq \slope{i}{j} \indNeuronPre{i}{j}\\
  &(\upperBound{\indNeuronPre{i}{j}} - \lowerBound{\indNeuronPre{i}{j}})\indNeuron{i}{j} - ( \upperBound{\indNeuronPre{i}{j}} \indNeuronPre{i}{j} - \upperBound{\indNeuronPre{i}{j}} \lowerBound{\indNeuronPre{i}{j}} ) \leq 0
\end{aligned}\right\} \\ &\quad\quad \forall i \in \{ 1, ..., \outLayerIdx-1 \}, \; j \in \reluUns_i\\
&\left.\begin{aligned}
  &\indNeuronTwoPost{i}{j} \geq \slopeHat{i}{j} \indNeuronTwoPre{i}{j}\\
  &(\upperBound{\indNeuronTwoPre{i}{j}} - \lowerBound{\indNeuronTwoPre{i}{j}})\indNeuronTwoPost{i}{j} - ( \upperBound{\indNeuronTwoPre{i}{j}} \hat{x}\uid{i} - \upperBound{\indNeuronTwoPre{i}{j}} \lowerBound{\indNeuronTwoPre{i}{j}} ) \leq 0
\end{aligned}\right\} \\ &\quad\quad \forall i \in \{ 1, ..., \outLayerIdx-1 \}, \; j \in \reluUnsHat_i\\
\end{aligned} 
\end{equation}}

and the constraints on \relu transformation of relational neurons derived from \diffpoly propagation rule (Table ~\ref{tab:diffpoly_1} and~\ref{tab:diffpoly_2}). When it comes to taking the case \exampleCase as an example, since $\relNeuronPost{i}{j} = \indNeuron{i}{j}$, the constraints are derived as:
{\mathcompact
\begin{equation}
\begin{aligned}
  &\relNeuronPost{i}{j} \geq \slope{i}{j} \indNeuronPre{i}{j},\quad  (\upperBound{\indNeuronPre{i}{j}} - \lowerBound{\indNeuronPre{i}{j}}) \relNeuronPost{i}{j} - (\upperBound{\indNeuronPre{i}{j}} \indNeuronPre{i}{j} - \upperBound{\indNeuronPre{i}{j}} \lowerBound{\indNeuronPre{i}{j}}) \leq 0 \\
  &\quad \quad \forall i \in \{ 1, ..., \outLayerIdx-1\}, \quad j \in \reluUns_i \cap \reluNegHat_i
\end{aligned}
\end{equation}}
$\slope{i}{j} \in [0,1]$ and $\slopeHat{i}{j} \in [0,1]$ denote the slope parameter controlling the lower bound of \relu~linear relaxation. The constraints for the remaining cases are given analogously. 

Since all \relu activations are replaced by linear relaxations, the primal problem is a linear program. Therefore, strong duality holds and the dual problem can be derived via the Karush-Kuhn-Tucker (\kkt) conditions.

\subsection{Lagrangian problem}\label{sec:lagrngian_prob}
Then, by utilizing objective and constraints, we can form the Lagrangian function. More concretely, we introduce Lagrangian multipliers to weight all constraints according to the stationary conditions and sum both objective and weighted constraints. The Lagrangian function \lagrFunc is expressed as:
{\mathcompact
\begin{equation}\label{eq:lagrangian_func}
\begin{aligned}
\lagrFunc = 
&C \cdot \relNeuronPost{\outLayerIdx}{}
+ \inpNeg \cdot (\inpSpaceLower{} - y\uid{0}) - \inpPos \cdot (\inpSpaceUpper{} - y\uid{0})\\
&+ \inpHatNeg \cdot (\inpSpaceLower{} - \hat{y}\uid{0}) - \inpHatPos \cdot (\inpSpaceUpper{} - \hat{y}\uid{0})\\
&- \inpDeltaNeg \cdot (\inpSpecDist + \relNeuronPost{0}{}) - \inpDeltaPos \cdot (\inpSpecDist - \relNeuronPost{0}{})\\
&+ \sum_{i=0}^{\outLayerIdx-1} \begin{aligned}\bigl(
  & -\affineInd[i+1] \cdot (W\uid{i}y\uid{i} + b\uid{i} - x\uid{i+1}) \\
  & - \affineIndHat[i+1] \cdot (W\uid{i}\hat{y}\uid{i} + b\uid{i} - \hat{x}\uid{i+1}) \\
  & - \affineDelta[i+1] \cdot (W\uid{i}\relNeuronPost{i}{} - \relNeuronPre{i+1}{}) \bigr)
\end{aligned}\\
& + \sum_{i=1}^{\outLayerIdx-1} \sum_{j} \left( \sum_{\constE \in \equC(i,j)} \dualVar(\constE) \constE + \sum_{\constI \in \ineC(i,j)} \dualVar(\constI) \constI \right)
\end{aligned}
\end{equation}}
where $\inpNeg, \inpPos, \inpHatNeg, \inpHatPos, \inpDeltaNeg, \inpDeltaPos \geq 0$ are Lagrangian multipliers for input inequality constraints and $\affineInd[i], \affineIndHat[i], \affineDelta[i]$ are Lagrangian multipliers for equality constraint of affine transformations.
In addition, $\equC(i,j)$ and $\ineC(i,j)$ are respectively sets of equality and inequality constraints for \relu transformation of \indNeuronOne{i}{j}. \constE denotes one of the equality constraints in $\equC(i,j)$, e.g. $\constE = 0$, while \constI denotes one of the inequality constraints in $\ineC(i,j)$, e.g. $\constI \leq 0$. $\dualVar(c)$ is a Lagrangian multiplier for the corresponding constraint $c$. Note that $\dualVar(ci)$ are imposed as non-negative (i.e. $\dualVar(ci) \geq 0$) on each inequality constraints, such as $ci \leq 0$, because of complementary slackness in the \kkt conditions. The sign of Lagrangian multipliers of input inequality constraints are also set by following the same rule. \kkt conditions are given according to that the primal problem has an optimal solution.

Same as the last example, we provide what the last term of Eq.~(\ref{eq:lagrangian_func}) becomes in the case \exampleCase.
{\mathcompact
\begin{equation}
\sum_{i=1}^{\outLayerIdx-1} \sum_{j \in \reluUns_i \cap \reluNegHat_i} \begin{aligned}\Bigl(
  &\reluIndUb[i]_j \bigl((\upperBound{\indNeuronPre{i}{j}} - \lowerBound{\indNeuronPre{i}{j}})\indNeuron{i}{j} - ( \upperBound{\indNeuronPre{i}{j}} x\uid{i} - \upperBound{\indNeuronPre{i}{j}} \lowerBound{\indNeuronPre{i}{j}} ) \bigr)\\
  &+ \reluIndLb{i}{j} (\slope{i}{j} \indNeuronPre{i}{j} - \indNeuron{i}{j}) 
  +\reluRelUnsNegLb{i}{j} (\slope{i}{j} \indNeuronPre{i}{j} - \relNeuronPost{i}{j})\\
  &+ \reluRelUnsNegUb[i]_j \bigl( (\upperBound{\indNeuronPre{i}{j}} - \lowerBound{\indNeuronPre{i}{j}}) \relNeuronPost{i}{j} - (\upperBound{\indNeuronPre{i}{j}} \indNeuronPre{i}{j} - \upperBound{\indNeuronPre{i}{j}} \lowerBound{\indNeuronPre{i}{j}}) \bigr) \Bigr)\end{aligned}
\end{equation}}
where $\reluIndUb[i]_j$ and $\reluRelUnsNegUb[i]_j$ are Lagrangian multipliers corresponding to the \relu linear relaxation constraints of upper bounds for $\indNeuron{i}{j}$ and $\relNeuronPost{i}{j}$, respectively. Similarly, $\reluIndLb{i}{j}$ and $\reluRelUnsNegLb{i}{j}$ are the lower ones for $\indNeuron{i}{j}$ and $\relNeuronPost{i}{j}$, respectively. Note that there actually exist terms regarding the equality constraints $\indNeuronTwo{i}{j} = 0$, but we omitted them because they (zero) do not affect.
The remaining cases are also expressed in the same way.

Consequently, the primal problem in \S{~\ref{sec:primal_prob}} is translated into the Lagrangian problem that is minimizing \lagrFunc with Lagrangian multipliers.

\subsection{Dual Problem}\label{sec:dual_prob}
The Lagrangian problem in \S{~\ref{sec:lagrngian_prob}} over the primal problem in \S{~\ref{sec:primal_prob}} can be converted into the Dual problem. Since the primal is an LP and strong duality holds, optimal primal and dual solutions satisfy the \kkt conditions. In particular, stationarity requires:
{\mathcompact
\begin{equation}
    \frac{\partial \lagrFunc}{\partial v} = 0
\end{equation}
}
for every primal  $v$ (i.e., $\indNeuronPre{i}{j}$, $\indNeuronTwoPre{i}{j}$, $\relNeuronPre{i}{j}$, $\indNeuronOne{i}{j}$, $\indNeuronTwo{i}{j}$, $\relNeuronPost{i}{j}$).

Since the final form is provided in \S{~\ref{sec:rel_neuron_selection}}, we show how we can reach it by taking the same case \exampleCase as an example.
The constant terms without relating to the variables corresponding to the case \exampleCase can be expressed by the summation of all such terms:
{\mathcompact
\begin{equation}
\sum_{i=1}^{\outLayerIdx-1} \sum_{j \in \reluUns_i \cap \reluNegHat_i} \begin{aligned}\bigl(
    & -\affineInd[i+1]_j b\uid{i}_j - \affineIndHat[i+1]_j b\uid{i}_j \\
    & + \reluIndUb[i]_j \upperBound{\indNeuronPre{i}{j}} \lowerBound{\indNeuronPre{i}{j}} + \reluRelUnsNegUb[i]_j \upperBound{\indNeuronPre{i}{j}} \lowerBound{\indNeuronPre{i}{j}}
\bigr)\end{aligned}
\end{equation}
}
On the other hand, the variable terms are derived from $\nabla \lagrFunc = 0$ over the variables:

{\mathcompact
\begin{align}
  &\affineIndHat[i]_j = \affineDelta[i]_j = 0 \label{eq:dual_prob_const1} \\
  &-\reluIndUb[i]_j \upperBound{\indNeuronPre{i}{j}} + \slope{i}{j} \reluIndLb{i}{j} + \slope{i}{j} \reluRelUnsNegLb{i}{j} - \reluRelUnsNegUb[i]_j \upperBound{\indNeuronPre{i}{j}} + \affineInd[i]_j = 0 \label{eq:dual_prob_const2} \\
  &\reluIndUb[i]_j (\upperBound{\indNeuronPre{i}{j}} - \lowerBound{\indNeuronPre{i}{j}}) - \reluIndLb{i}{j} - (W^{(i)}\transposed \affineInd[i+1])_j = 0 \label{eq:dual_prob_const3} \\
  &-\reluRelUnsNegLb{i}{j} + \reluRelUnsNegUb[i]_j (\upperBound{\indNeuronPre{i}{j}} - \lowerBound{\indNeuronPre{i}{j}}) - (W^{(i)}\transposed \affineDelta[i+1])_j = 0 \label{eq:dual_prob_const4} \\
  & \quad \quad \forall i \in \{1, ..., \outLayerIdx-1\}, \quad j \in \reluUns_i \cap \reluNegHat_i \notag
\end{align}
}
The top line (\ref{eq:dual_prob_const1}) is given from $\partial \lagrFunc / \partial \indNeuronTwoPre{i}{j} = \partial \lagrFunc / \partial \relNeuronPre{i}{j} = 0$, the second line (\ref{eq:dual_prob_const2}) is derived from $\partial \lagrFunc / \partial \indNeuronPre{i}{j} = 0$, and the third and forth line (\ref{eq:dual_prob_const3}, \ref{eq:dual_prob_const4}) are respectively derived from $\partial \lagrFunc / \partial \indNeuron{i}{j} = 0$ and $\partial \lagrFunc / \partial \relNeuronPost{i}{j} = 0$.

By taking the \kkt conditions into the consideration, Lagrangian multipliers (they are also called Dual variables in Dual context) behave as indicators of constraint tightness. In other words, they become zero whenever the corresponding constraints are inactive, and become non-zero only when they are active. Since the upper and lower bounds of \relu relaxation constraints cannot be simultaneously active, complementary slackness implies that at most one of the corresponding dual variables is non-zero, i.e., either $\reluIndUb[i]_j$ or $\reluIndLb{i}{j}$ is zero. By following this rule and non-negativity of $\reluIndUb[i]_j, \reluIndLb{i}{j}, \reluRelUnsNegUb[i]_j, \reluRelUnsNegLb{i}{j}$, we obtain sign-based decomposition from third and fourth line (\ref{eq:dual_prob_const3}, \ref{eq:dual_prob_const4}):
{\mathcompact
\begin{equation}\label{eq:sign_based_decomp}
\begin{aligned}
    & \reluIndUb[i]_j (\upperBound{\indNeuronPre{i}{j}} - \lowerBound{\indNeuronPre{i}{j}}) = [(W^{(i)}\transposed \affineInd[i+1])_j]_+ \\
    & \reluIndLb{i}{j} = [(W^{(i)}\transposed \affineInd[i+1])_j]_- \\
    & \reluRelUnsNegUb[i]_j (\upperBound{\indNeuronPre{i}{j}} - \lowerBound{\indNeuronPre{i}{j}}) = [(W^{(i)}\transposed \affineDelta[i+1])_j]_+ \\
    & \reluRelUnsNegLb{i}{j} = [(W^{(i)}\transposed \affineDelta[i+1])_j]_-
\end{aligned}
\end{equation}
}
where $[z]_- = \max(0,-z)$ and $[z]_+ = \max(0,z)$. In addition, by substituting Eq.~(\ref{eq:sign_based_decomp}) into the second line~(\ref{eq:dual_prob_const2}) and using $\preReluInd[i]_j$ and $\preReluDelta[i]_j$ as a shorthand for $(W^{(i)}\transposed \affineInd[i+1])_j$ and $(W^{(i)}\transposed \affineDelta[i+1])_j$, we obtain:
{\mathcompact
\begin{equation}\label{eq:dual_prob_A}
    \affineInd[i]_j = \frac{\upperBound{\indNeuronPre{i}{j}}}{\upperBound{\indNeuronPre{i}{j}} - \lowerBound{\indNeuronPre{i}{j}}} ([\preReluInd[i]_j]_+ + [\preReluDelta[i]_j]_+) 
    - \slope{i}{j} ([\preReluInd[i]_j]_- + [\preReluDelta[i]_j]_-)
\end{equation}
}
This equation shows that the dual variables propagate backward (e.g., $\affineDelta[i+1])_j$ to $\affineDelta[i])_j$) through the network in a piecewise-linear manner, where the coefficients depend on the \relu relaxation bounds and the sign of the backpropagated quantities.
For the remaining cases and terms regarding inputs, we can derive the transformation as shown in Eq.~(\ref{eq:sign_based_decomp}) and (\ref{eq:dual_prob_A}).
In practice, we set $\slope{i}{j} = \frac{\upperBound{\indNeuronPre{i}{j}}}{\upperBound{\indNeuronPre{i}{j}} - \lowerBound{\indNeuronPre{i}{j}}}$ and $\slopeHat{i}{j} =  \frac{\upperBound{\indNeuronTwoPre{i}{j}}}{\upperBound{\indNeuronTwoPre{i}{j}} - \lowerBound{\indNeuronTwoPre{i}{j}}}$. With these choices, the \relu relaxation is handled by two parallel linear bounds. 

Therefore, by eliminating primal variables (e.g., $\indNeuron{i}{j}$, $\relNeuronPre{i}{j}$) using the stationary conditions and substituting the sign-based decomposition of dual variables, we obtain a backward propagation rule for dual variables. Applying this procedure layer by layer yields the the dual formulation in \S{\ref{sec:primal_prob}}.

\end{document}